%% file: acl2021.tex
\pdfoutput=1
\documentclass[11pt]{article}

\usepackage{authblk}
\usepackage[hyperref]{acl2021}
\usepackage{times}
\usepackage{latexsym}
\usepackage{amsmath}
\usepackage{cleveref}
\usepackage{subfiles} 
\usepackage{amssymb}
\usepackage{booktabs}
\usepackage[utf8x]{inputenc}
\usepackage{url}
\usepackage{multirow}
\usepackage{array}
\usepackage{graphicx} 
\usepackage{textcomp}
\usepackage{subcaption} % do not add any optional arguments!
\usepackage[inline]{enumitem}
\usepackage{bm} 
\usepackage{bold-extra} 
\usepackage{empheq}   
\usepackage{amssymb} 
\usepackage{mathtools}
\usepackage{pifont}
\usepackage{enumitem}
\usepackage{float}
\usepackage{stfloats}
\usepackage{numprint}

\newcolumntype{L}[1]{>{\raggedright\arraybackslash} m{#1\textwidth}}
\newcolumntype{R}[1]{>{\raggedleft\arraybackslash} m{#1\textwidth}}
\newcolumntype{C}[1]{>{\centering\arraybackslash} m{#1\textwidth}}
\newcolumntype{X}[1]{>{\raggedright\arraybackslash} m{#1\columnwidth}}
\newcolumntype{Y}[1]{>{\raggedleft\arraybackslash} m{#1\columnwidth}}
\newcolumntype{Z}[1]{>{\centering\arraybackslash} m{#1\columnwidth}}

\aclfinalcopy % Uncomment this line for the final submission
\def\aclpaperid{288} %  Enter the Paper ID here

\usepackage{microtype}

\newcommand{\onlyinsubfile}[1]{#1}
\newcommand{\notinsubfile}[1]{}
\newcommand{\myeq}{{\mkern1.0mu{=}\mkern1.0mu}}
\newcommand{\best}[1]{\textbf{{#1}}}
\newcommand{\insep}[1]{\textbf{{#1}}}

\newcommand{\mn}{namysl-etal-2020-nat}
\newcommand{\mnp}{\citep{\mn}}
\newcommand{\mnt}{\citet{\mn}}
\newcommand{\seq}{seq2seq}
\newcommand{\ch}{sentence-level}
\newcommand{\tok}{token-level}
\newcommand{\charseq}{{\seq} {({\ch}, \S\ref{ssec:sent_tok_modeling})}}
\newcommand{\tokseq}{{\seq} {({\tok}, \S\ref{ssec:sent_tok_modeling})}}
\newcommand{\natast}{{\citet{hamalainen-hengchen-2019-paft}}}
\newcommand{\meanstddev}[2]{{#1$\pm$#2}}
\newcommand{\cmark}{\text{\ding{51}}}
\newcommand{\xmark}{\text{\ding{55}}}
\newcommand{\given}[1]{{\mkern1.0mu\vert\mkern1.0mu#1}}
\newcommand{\na}{n/a}
\newcommand{\none}{---}

\newcommand\narrowstyleA[1]{{\SetTracking{encoding=*}{-4}\lsstyle#1}}
\newcommand\narrowstyleB[1]{{\SetTracking{encoding=*}{-9}\lsstyle#1}}
\newcommand\narrowstyleC[1]{{\SetTracking{encoding=*}{-15}\lsstyle#1}}
\newcommand\narrowstyleD[1]{{\SetTracking{encoding=*}{-20}\lsstyle#1}}

\title{Empirical Error Modeling Improves Robustness of Noisy Neural Sequence Labeling}

\author[1,2]{{\bf Marcin Namysl}}
\author[1,2]{{\bf Sven Behnke}}
\author[1]{{\bf Joachim K\"ohler}}

\affil[1]{
Fraunhofer IAIS\\
Sankt Augustin, Germany
}

\affil[2]{
Autonomous Intelligent Systems\\
University of Bonn, Germany
}

\affil[ ]{
\texttt{\{Marcin.Namysl,Sven.Behnke,Joachim.Koehler\}@iais.fraunhofer.de}
}

\date{}

\begin{document}

\renewcommand{\onlyinsubfile}[1]{}
\renewcommand{\notinsubfile}[1]{#1}

\maketitle

\begin{abstract}
Despite recent advances, standard sequence labeling systems often fail when processing noisy user-generated text or consuming the output of an Optical Character Recognition (OCR) process.
In this paper, we improve the noise-aware training method by proposing an empirical error generation approach that employs a sequence-to-sequence model trained to perform translation from error-free to erroneous text.
Using an OCR engine, we generated a large parallel text corpus for training and produced several real-world noisy sequence labeling benchmarks for evaluation.
Moreover, to overcome the data sparsity problem that exacerbates in the case of imperfect textual input, we learned noisy language model-based embeddings.
Our approach outperformed the baseline noise generation and error correction techniques on the erroneous sequence labeling data sets.
To facilitate future research on robustness, we make our code, embeddings, and data conversion scripts publicly available.
\end{abstract}

\subfile{sections/introduction}

\subfile{sections/related_work}

\subfile{sections/problem_definition}

\subfile{sections/proposed_method}

\subfile{sections/experiments}

\subfile{sections/conclusions}

\bibliography{../anthology,../acl2021}
\bibliographystyle{../acl_natbib}

\appendix

\ifaclfinal

\subfile{sections/appendix}

\fi

\end{document}

%% file: sections/introduction.tex
\makeatletter
\let\maintitle\@title
\title{
\maintitle\\
(Introduction)
}
\makeatother

\maketitle

\section{Introduction}
\label{sec:intro}

Deep learning models have already surpassed human-level performance in many Natural Language Processing (NLP) tasks\footnote{GLUE benchmark~\citep{wang-etal-2018-glue}: \url{https://gluebenchmark.com/leaderboard}}.
Sequence labeling systems have also reached extremely high accuracy~\citep{akbik-etal-2019-pooled,heinzerling-strube-2019-sequence}.
Still, NLP models often fail in scenarios, where non-standard text is given as input~\citep{heigold-etal-2018-robust,DBLP:conf/iclr/BelinkovB18}.

NLP algorithms are predominantly trained on error-free textual data but are also employed to process user-generated text~\citep{baldwin-etal-2013-noisy,derczynski-etal-2013-twitter} or consume the output of prior Optical Character Recognition (OCR) or Automatic Speech Recognition (ASR) processes~\citep{miller-etal-2000-named}.
Errors that occur in any upstream component of an NLP system deteriorate the accuracy of the target downstream task~\citep{Alex:2014:ERQ:2595188.2595214}.

In this paper, we focus on the problem of performing sequence labeling on the text produced by an OCR engine. 
Moreover, we study the transferability of the methods learned to model OCR noise to the distribution of the human-generated errors.
Both misrecognized and mistyped text pose a challenge for the standard models trained using error-free data~\citep{\mn}.

\begin{figure}[!t]
\begin{center}
\includegraphics[width=1.0\columnwidth]{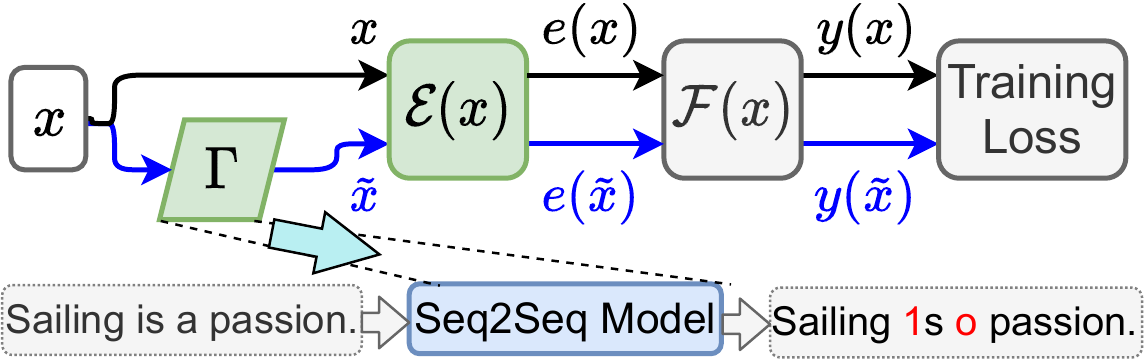}
\caption{
\narrowstyleA
{
Our modification of the NAT approach (green boxes).
We propose a learnable seq2seq-based error generator and re-train FLAIR embeddings using noisy text to improve the accuracy of noisy neural sequence labeling.
$\Gamma$ is a process that induces noise to the input $x$ producing erroneous $\tilde{x}$.
$\mathcal{E}(x)$ is an embedding matrix.
$\mathcal{F}(x)$ is a sequence labeling model.
$e(x)$ and $e(\tilde{x})$ are the embeddings of $x$ and $\tilde{x}$, respectively.
$y(x)$ and $y(\tilde{x})$ are the outputs of the model for $x$ and $\tilde{x}$, respectively. 
}}
\label{fig:train_arch}
\end{center}
\end{figure}

We make the following contributions (\Cref{fig:train_arch}):
\begin{itemize}
\item We propose a noise generation method for OCR that employs a sequence-to-sequence (seq2seq) model trained to translate from error-free to erroneous text (\S\ref{ssec:noise_generation}).
Our approach improves the accuracy of noisy neural sequence labeling compared to prior work (\S\ref{ssec:eval-err-gen}).
\item We present an unsupervised parallel training data generation method that utilizes an OCR engine (\S\ref{ssec:data_generation}).
Similarly, realistic noisy versions of popular sequence labeling data sets can be synthesized for evaluation (\S\ref{ssec:noisy_datasets}).
\item We exploit erroneous text to perform Noisy Language Modeling (NLM; \S\ref{ssec:noisy_lm}). Our NLM embeddings further improve the accuracy of noisy neural sequence labeling (\S\ref{ssec:eval-noisy-lm}), also in the case of the human-generated errors (\S\ref{ssec:eval-typos}).
\item To facilitate future research on robustness, we integrate our methods into the Noise-Aware Training (NAT) framework {\mnp} and make our code, embeddings, and data conversion scripts publicly available.\footnote{\narrowstyleB{{\url{https://github.com/mnamysl/nat-acl2021}}}}
\end{itemize}

\onlyinsubfile{
\bibliographystyle{../acl_natbib}
\bibliography{../anthology,../acl2021}
}

%% file: sections/related_work.tex
\makeatletter
\let\maintitle\@title
\title{
\maintitle\\
(Related Work)
}
\makeatother

\maketitle

\section{Related Work}
\label{sec:related_work}

Errors of OCR, ASR, and other text generators always pose a challenge to the downstream NLP systems~\citep{Lopresti2009,10.1145/1871840.1871845,DBLP:conf/interspeech/RuizGBF17}.
Hence, methods for improving robustness are becoming increasingly popular.

%=================== Data augmentation ===================

\paragraph{Data Augmentation}
A widely adopted method of providing robustness to non-standard input is to augment the training data with examples perturbed using a model that mimics the error distribution to be encountered at test time~\citep{Cubuk_2019_CVPR}. 

Apparently, the exact modeling of noise might be impractical or even impossible---thus, methods that employ randomized error patterns for training recently gained increasing popularity~\citep{heigold-etal-2018-robust,lakshmi-narayan-etal-2019-exploration}.
Although trained using synthetic errors, these methods are often able to achieve moderate improvements on data from natural sources of noise~\citep{DBLP:conf/iclr/BelinkovB18,karpukhin-etal-2019-training}.

%=================== Spelling- and OCR Post-Correction ===================

\paragraph{Spelling- and OCR Post-correction}

The most widely used method of handling noisy text is to apply error correction on the input produced by human writers (\textit{spelling correction}) or the output of an upstream OCR component (\textit{OCR post-correction}).

A popular approach applies \textit{monotone seq2seq} modeling for the correction task~\citep{schnober-etal-2016-still}. 
For instance, 
\natast~proposed \textit{Natas}---an OCR post-correction method that uses character-level Neural Machine Translation (NMT).
They extracted parallel training data using embeddings learned from the erroneous text and used it as input to their translation model.

%=================== Grammatical Error Correction ===================

\paragraph{Grammatical Error Correction}

Grammatical Error Correction (GEC;~\citealp{ng-etal-2013-conll,ng-etal-2014-conll,bryant-etal-2019-bea}) aims to automatically correct ungrammatical text.
GEC can be approached as a translation from an ungrammatical to a grammatical language, which enabled NMT seq2seq models to be applied to this task~\cite{yuan-briscoe-2016-grammatical}.
Due to the limited size of human-annotated GEC corpora, NMT models could not be trained effectively~\cite{lichtarge-etal-2019-corpora}, though.

%
% Artificial Error Generation (AEG)
%
Several studies investigated generating realistic erroneous sentences from grammatically correct text to boost training data~\cite{kasewa-etal-2018-wronging,grundkiewicz-etal-2019-neural,choe-etal-2019-neural,qiu-park-2019-artificial}.
Inspired by \emph{back-translation}~\cite{sennrich-etal-2016-improving,edunov-etal-2018-understanding}, Artificial Error Generation (AEG) approaches~\cite{rei-etal-2017-artificial,xie-etal-2018-noising} train an intermediate model in reverse order---to translate correct sentences to erroneous ones.
Following AEG, we generate a large corpus of clean and noisy sentences and train a seq2seq model to produce rich and diverse errors resembling the natural noise distribution (\S\ref{ssec:realistic-error-modeling},~\ref{ssec:data_generation}).

%=================== Noise-invariant latent representations ===================

\paragraph{Noise-Invariant Latent Representations}

Robustness can also be improved by encouraging the models to learn a similar latent representation for both the error-free and the erroneous input.

\citet{DBLP:conf/cvpr/ZhengSLG16} introduced \textit{stability training}---a general method used to stabilize predictions against small input perturbations.
\citet{piktus-etal-2019-misspelling} proposed \textit{Misspelling Oblivious Embeddings} that embed the misspelled words close to their error-free counterparts.
\citet{jones-etal-2020-robust} developed \emph{robust encodings} that balance stability (consistent predictions across various perturbations) and fidelity (accuracy on unperturbed input) by mapping sentences to a smaller discrete space of encodings.
Although their model improved robustness against small perturbations, it decreased accuracy on the error-free input.

%=================== Noise-Aware Training ===================

Recently, {\mnt}~proposed the Noise-Aware Training method that employs stability training and data augmentation objectives.
They exploited both the error-free and the noisy samples for training and used a \emph{confusion matrix-based} error model to imitate the errors.
In contrast to their approach, we employ a more realistic empirical error distribution during training (\S\ref{ssec:realistic-error-modeling}) and observe improved accuracy at test time (\S\ref{ssec:eval-err-gen}).

\onlyinsubfile{
\bibliographystyle{../acl_natbib}
\bibliography{../anthology,../acl2021}
}

%% file: sections/problem_definition.tex
\makeatletter
\let\maintitle\@title
\title{
\maintitle\\
(Problem Definition)
}
\makeatother

\maketitle

\section{Problem Definition}
\label{sec:problem}

\subsection{Noisy Neural Sequence Labeling}
\label{ssec:noisy-seq-lab}

{\mnt} pointed out that the standard NLP systems are generally trained using error-free textual input, which causes a discrepancy between the training and the test conditions.
These systems are thus more susceptible to non-standard, corrupted, or adversarial input.

To model this phenomenon, they formulated the \emph{noisy neural sequence labeling} problem, assuming that every input sentence might be subjected to some unknown token-level noising process $\Gamma\myeq P(\tilde{x}_i\given{x_i})$, where $x_i$ is the original $i$-th token, and $\tilde{x}_i$ is its distorted equivalent.
As a solution, they proposed the NAT framework, which trains the sequence labeling model using auxiliary objectives that exploit both the original sentences and their copies corrupted using a noising process that imitates the naturally occurring errors (\Cref{fig:train_arch}).

\subsection{Confusion Matrix-Based Error Model}
\label{ssec:cmx-model}

{\mnt} used a confusion matrix-based method to model insertions, deletions, and substitutions of characters. 
Given a corpus of paired noisy and manually corrected sentences $\mathcal{P}$, they estimated the natural error distribution by calculating the alignments between the pairs $(\tilde{x},x)\in\mathcal{P}$ of noisy and clean sentences using the \emph{Levenshtein distance} metric~\citep{Levenshtein1966a}.

Moreover, as $\mathcal{P}$ is usually laborious to obtain, they proposed a \emph{vanilla error model}, which assumes that all types of edit operations are equally likely:
\begin{align*}
\smashoperator[lr]{\sum_{\tilde{c}\,\in\,\Sigma{\setminus}\{\varepsilon\}}}P_{ins}(\tilde{c}\given{\varepsilon})
=
P_{del}(\varepsilon\given{c})
=
\smashoperator[lr]{\sum_{\tilde{c}\,\in\,\Sigma{\setminus}\{c,\,\varepsilon\}}}P_{subst}(\tilde{c}\given{c}),
\end{align*}
\noindent
where $c$ and $\tilde{c}$ are the original and the perturbed characters, respectively, $\Sigma$ is an alphabet, and $\varepsilon$ is a symbol introduced to model insertion and deletions.

\subsection{Realistic Empirical Error Modeling}
\label{ssec:realistic-error-modeling}

{\mnt}~compared the NAT models that used the vanilla- and the empirically-estimated confusion matrix-based error model and observed no advantages of exploiting the test-time error distribution during training.
\emph{Would we make the same observation given a more realistic error model?}

Even though the methods that used randomized error patterns were often successful, we argue that leveraging the empirical noise distribution for training would be beneficial, providing additional accuracy improvements.
The data produced by the na{\"\i}ve noise generation methods may not resemble naturally occurring errors, which could lead the downstream models to learn misleading patterns.

\begin{figure}[!ht]
\begin{center}
\includegraphics[width=.95\columnwidth]{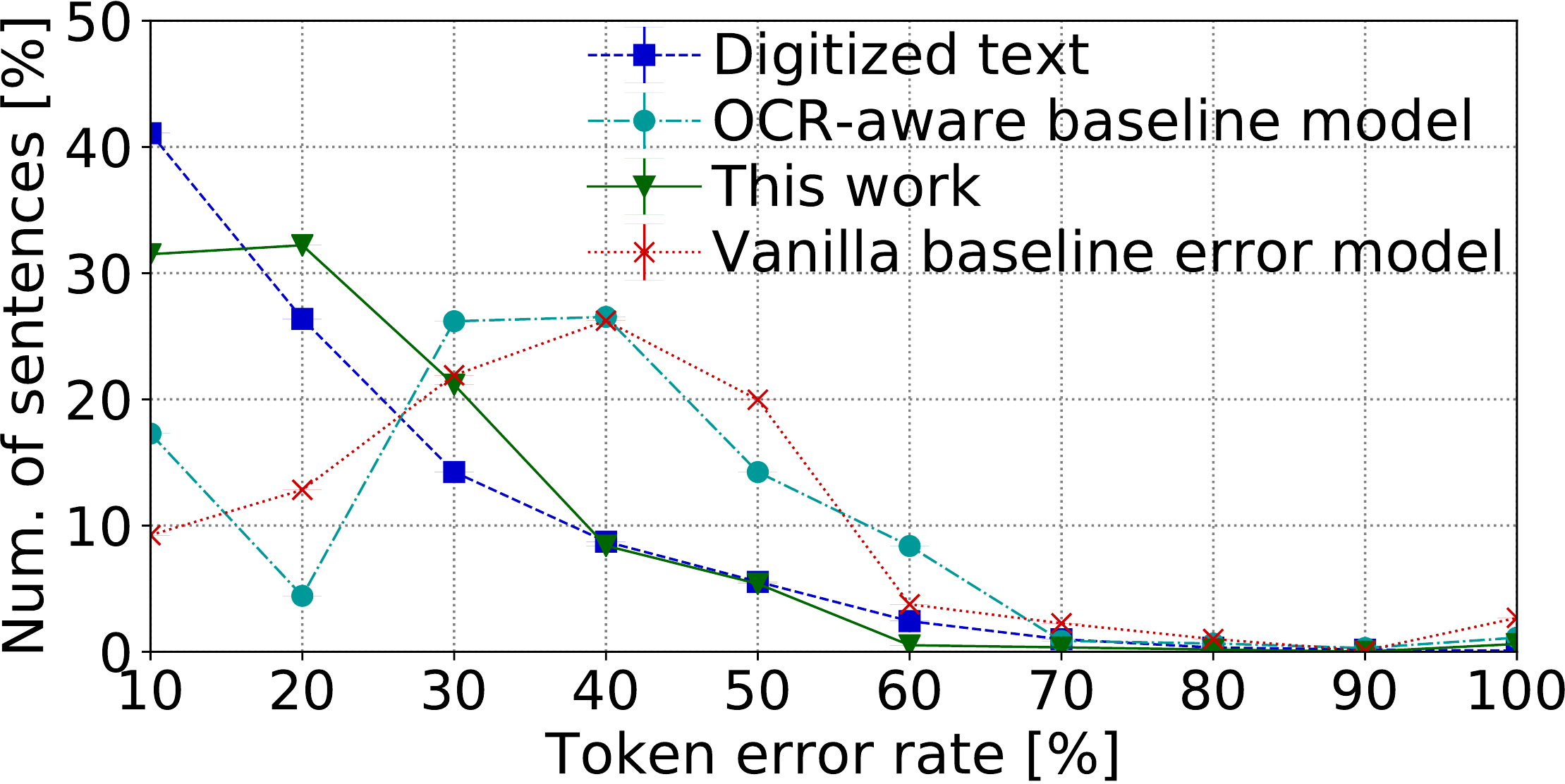}
\caption{
Distributions of the token error rates of sentences produced by the proposed and the baseline error models.
For comparison, we plot the distribution of error rates in the text that contains naturally occurring errors.
Each value $n$ is the percentage of sentences with a token error rate in $[n-10,n)$. 
}
\label{fig:error-distrib}
\end{center}
\end{figure}

In \Cref{fig:error-distrib}, we compare the distributions of error rates of sentences produced by the proposed and the prior noise models with the distribution of errors in the digitized text. 
We can observe that the distribution of naturally occurring errors follows Zipf's law, while the baseline noise models produce Bell-shaped curves.
Interestingly, both the vanilla and the empirical models exhibit similar characteristics, which could explain the observations from the prior work. 
In practice, the error rate is not uniform throughout the text.
Some passages are recognized perfectly, while others can barely be deciphered.
Our objective is thus to develop a noise model that produces a smoother distribution, imitating the errors encountered at test time more precisely (cf. \emph{This work} in \Cref{fig:error-distrib}).

Moreover, although the exact noise distribution in the test data cannot always be known beforehand, the noising process, e.g., an OCR engine, used to provide the input, can often be identified.
We would thus take advantage of such prior knowledge to improve the efficiency of the downstream task.

\subsection{Data Sparsity of Natural Language}
\label{ssec:data-sparsity}

\emph{Embeddings} pre-trained on a large corpus of monolingual text are ubiquitous in NLP~\cite{10.5555/2999792.2999959,peters-etal-2018-deep,devlin-etal-2019-bert}.
They capture syntactic and semantic textual features that can be exploited to solve higher-level NLP tasks.

Embeddings are generally trained using corpora that contain error-free text.
Due to the data sparsity problem that arises from the large vocabulary sizes and the exponential number of feasible contexts, the majority of possible word sequences do not appear in the input data.
Even though increasing the size of the training corpora was shown to improve the performance of language processing tasks~\cite{brown2020language}, most of the misrecognized or mistyped tokens would still be unobserved and therefore poorly modeled when using the error-free text only.
\emph{Would it be beneficial to pre-train the embeddings on data that includes realistic erroneous sentences?}

\subsection{The Flaws of Error Correction}

Furthermore, we believe that the correction methods, although widely adopted, can only reliably manage moderately perturbed text~\cite{flor-etal-2019-benchmark}.
OCR post-correction has been reported to be challenging in the case of historical books that exhibit high OCR error rates~\citep{8978127}.

We note that correction methods have no information about the downstream task to be performed.
Moreover, in the automatic correction setting, they only provide the best guess for each token.
Comparing their performance with the NAT approach in the context of sequence labeling would be informative.

\onlyinsubfile{
\bibliographystyle{../acl_natbib}
\bibliography{../anthology,../acl2021}
}

%% file: sections/proposed_method.tex
\makeatletter
\let\maintitle\@title
\title{
\maintitle\\
(Proposed Method)
}
\makeatother

\maketitle

\section{Empirical Error Modeling}
\label{sec:proposed_method}

\Cref{fig:train_arch} presents our modifications of the NAT framework. 
Firstly, we propose to replace the confusion matrix-based noising process (\S\ref{ssec:cmx-model}) with a noise induction method that generates a more realistic error distribution (\S\ref{ssec:noise_generation}-\ref{ssec:data_alignment}).
Secondly, to overcome the data sparsity problem (\S\ref{ssec:data-sparsity}), we train language model-based embeddings using digitized text and use them as a substitution of the pre-trained model used in prior work (\S\ref{ssec:noisy_lm}).

%==================== Sequence-to-Sequence Error Generator ====================

\subsection{Sequence-to-Sequence Error Generator}
\label{ssec:noise_generation}

Motivated by the AEG approaches~\cite{rei-etal-2017-artificial,xie-etal-2018-noising}, we propose a learnable error generation method that employs a character-level seq2seq model to perform monotone string translation~\citep{schnober-etal-2016-still}.
It directly models the conditional probability $p(\tilde{x}\given{x})$ of mapping error-free text $x$ into erroneous text $\tilde{x}$ using an attention-based encoder-decoder framework~\citep{38ed090f8de94fb3b0b46b86f9133623}.
The encoder computes the representation $h\myeq\{h_1,\ldots,h_{n}\}$ of $x$, where $n$ is the length of $x$. 
The decoder generates $\tilde{x}$ one token at a time:
\begin{align*}
p(\tilde{x}\given{x}) = \prod\nolimits_{i=1}^{n} p(\tilde{x}_{i}\given{\tilde{x}_{<i},x,c}),
\end{align*}
where $c\myeq f_{attn}(\{h_1,\ldots,h_{n}\})$ is a vector generated from $h$, and $f_{attn}$ is an attention function.

Our models are trained to maximize the likelihood of the training data.
At inference time, we randomly sample the subsequent tokens from the learned conditional language model.

\begin{figure}[!h]
\begin{center}
\includegraphics[width=.95\columnwidth]{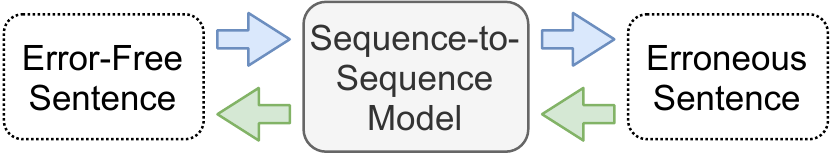}
\caption{
Schematic visualization of the error generation (blue arrows) and the error correction (green arrows) methods.
The parallel data can be utilized to train seq2seq models for both tasks.
}
\label{fig:error_correction_and_generation}
\end{center}
\end{figure}

Note that our approach reverses the standard seq2seq error correction pipeline, which uses the erroneous text as input and trains the model to produce the corresponding error-free string (\Cref{fig:error_correction_and_generation}).
By interchanging the input and the output data, we can also readily train sentence correction models. 
One difference is that at inference time we would prefer to perform beam search and select the best decoding result rather than sampling subsequent characters from the learned distribution.

%==================== Unsupervised Parallel Data Generation ====================

\subsection{Unsupervised Parallel Data Generation}
\label{ssec:data_generation}

To train our error generation model (\S\ref{ssec:noise_generation}), we need a large parallel corpus $\mathcal{P}$ of error-free and erroneous sentences. 
AEG approaches use seed GEC corpora to learn the inverse models directly. 
Unfortunately, we are not aware of any comparably large resources for digitized text that could be used for this task. 

\begin{figure}[!htb]
\begin{center}
\includegraphics[width=0.96\columnwidth]{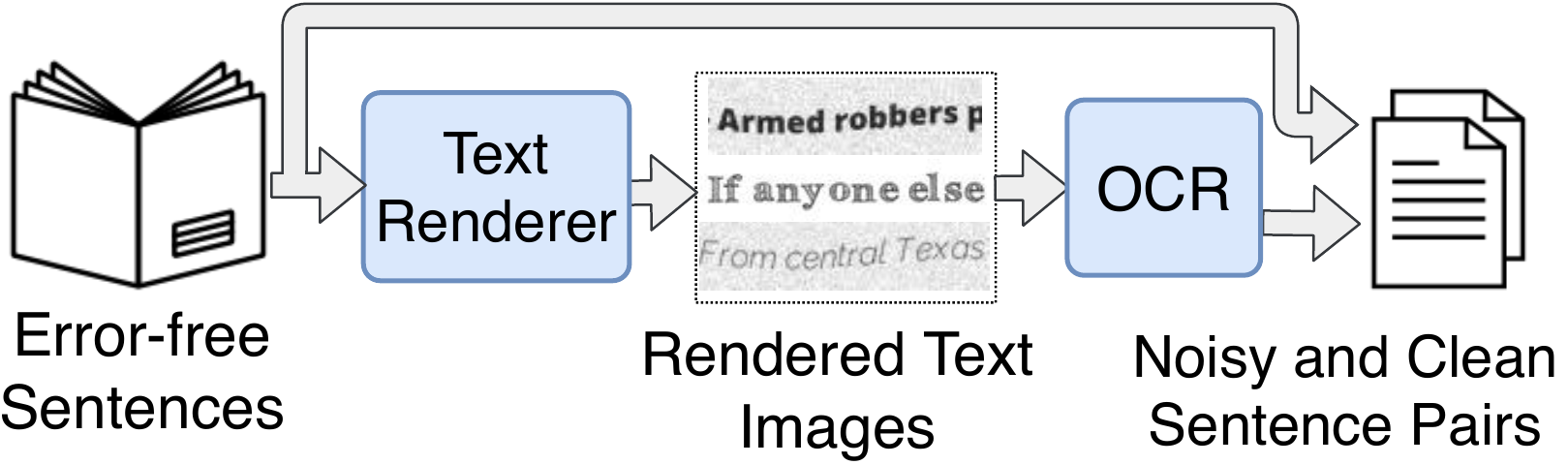}
\caption{
\narrowstyleA
{
Our parallel data generation method for OCR.
We render sentences extracted from a text corpus.
Subsequently, an OCR engine recognizes the text depicted in the rendered images.
Finally, the pairs of original and recognized sentences are gathered together to form a parallel corpus used to train translation models.
}}
\label{fig:data_generation}
\end{center}
\end{figure}

To address this issue, we propose an unsupervised sentence-level parallel data generation approach for OCR (\Cref{fig:data_generation}).
First, we collect a large seed corpus $\mathcal{T}$ that contains the error-free text.
We then render each sentence and subsequently run text recognition on the rendered images using an OCR engine. 
Moreover, to increase the variation in training data, we sample different fonts for rendering.
Furthermore, to simulate the distortions and degradation of the printed material, we induce pixel-level noise to the images before recognition.

Note that our approach is universal and could be used to generate parallel data sets for other tasks, e.g., an ASR system could be trained on samples from a Text-to-Speech engine~\cite{wang-etal-2018-hybrid}.

%==================== Sentence- and Word-level Modeling ====================

\subsection{Sentence- and Word-Level Modeling}
\label{ssec:sent_tok_modeling}

We note that the sequence labeling problem is formulated at the word-level, i.e., each word has a class label assigned to it.
To employ our method in this scenario, we develop
\begin{enumerate*}[label=(\roman*)]
\item a \emph{\ch} and
\item a \emph{\tok}
\end{enumerate*}
variant of our error generator.

Our {\ch} error generator uses a seq2seq model trained to translate from error-free to erroneous sentences.
It can potentially utilize contextual information from surrounding words, which may improve the quality of the results.
During the training of a NAT model, a learned seq2seq model translates the original input $x$ to generate $\tilde{x}$. 
Subsequently, we use an alignment algorithm (\S\ref{ssec:data_alignment}) to transfer the word-level annotations from $x$ to $\tilde{x}$. 

Our {\tok} error generator uses a seq2seq model trained to translate from error-free to erroneous words. 
It relies exclusively on the input and the output words.
We use the alignment algorithm to build a training set for this task, i.e., extract word-level parallel data from the 
corpus of parallel sentences (\S\ref{ssec:data_generation}).
During the training of a NAT model, a learned generator translates each word $x_{i}$ from $x$ to produce the erroneous sentence $\tilde{x}$. 

%==================== Word-Level Sentence Alignment ====================

\subsection{Word-Level Sentence Alignment}
\label{ssec:data_alignment}

\begin{figure}[!htbp]
\begin{center}
\includegraphics[width=1.0\columnwidth]{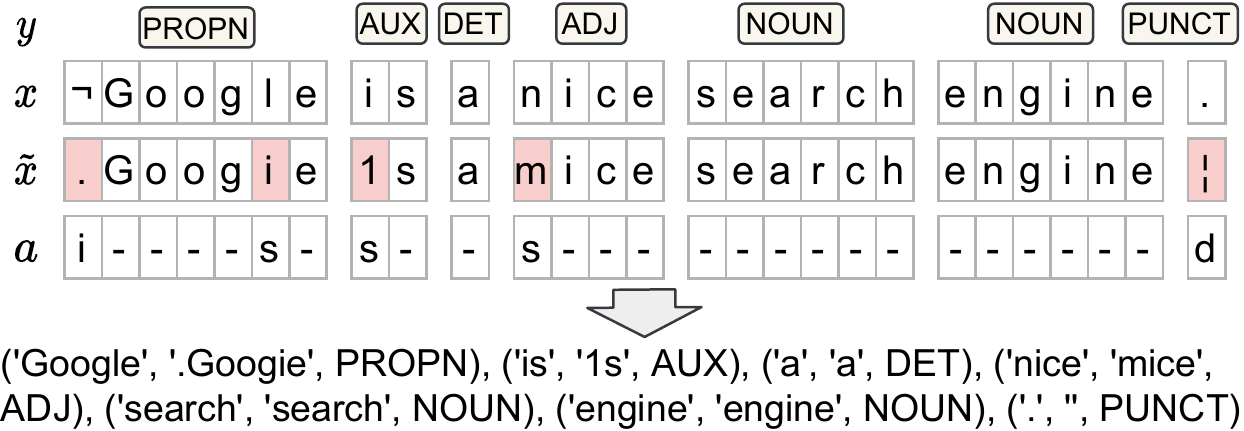}
\caption{
\narrowstyleC
{
Our sentence alignment procedure.
We align the original and the recognized sentences ($x$ and $\tilde{x}$, respectively) using the sequence of edit operations $a$, which include insertions "{i}", deletions "{d}", and substitutions "{s}" of characters.
We use "\unichar{"00AC}" and "\unichar{"00A6}" as placeholders for the insertion and the deletion operation, respectively. 
Matched characters are marked with "-". The alignment procedure produces a list of paired error-free and possibly erroneous words with class labels (optional).
}}
\label{fig:sentence_alignment}
\end{center}
\end{figure}

\Cref{fig:sentence_alignment} illustrates the alignment procedure, which we developed to extract word-level parallel training data for our {\tok} generator and to transfer the labels to the erroneous sentences for the {\ch} generator in the sequence labeling scenario.

To this end, we align each pair of error-free and noisy sentences at the word-level using the Levenshtein Distance algorithm.
Our alignment procedure produces pairs of aligned words.
The annotations for words are transferred accordingly.

%

%==================== Noisy Language Modeling ====================

\subsection{Noisy Language Modeling}
\label{ssec:noisy_lm}

Recently, \citet{noising2017} drew a connection between input noising in neural network language models and smoothing in n-gram models. 
We believe that data noising could be an effective technique for regularizing neural language models that could help to overcome the data sparsity problem of imperfect natural language text and enable learning meaningful representation of erroneous tokens.

To this end, we propose to include the data from noisy sources in the corpora used to train LM-based embeddings.
Specifically, in this work, we learn a noisy language model using the output of an OCR engine (\S\ref{ssec:data_generation}) that captures the characteristics of OCR errors.
Any other noisy source could be readily used to model related domains, e.g., ASR-transcripts or ungrammatical text.

\onlyinsubfile{
\bibliographystyle{../acl_natbib}
\bibliography{../anthology,../acl2021}
}

%% file: sections/experiments.tex
\makeatletter
\let\maintitle\@title
\title{
\maintitle~(Evaluation)
}
\makeatother

\maketitle

\section{Experimental Setup}
\label{sec:setup}

%==================== Sequence-to-Sequence Modeling ====================

\subsection{Sequence-to-Sequence Error Generator}

\begin{figure}[!htb]
\begin{center}
\includegraphics[width=1.\columnwidth]{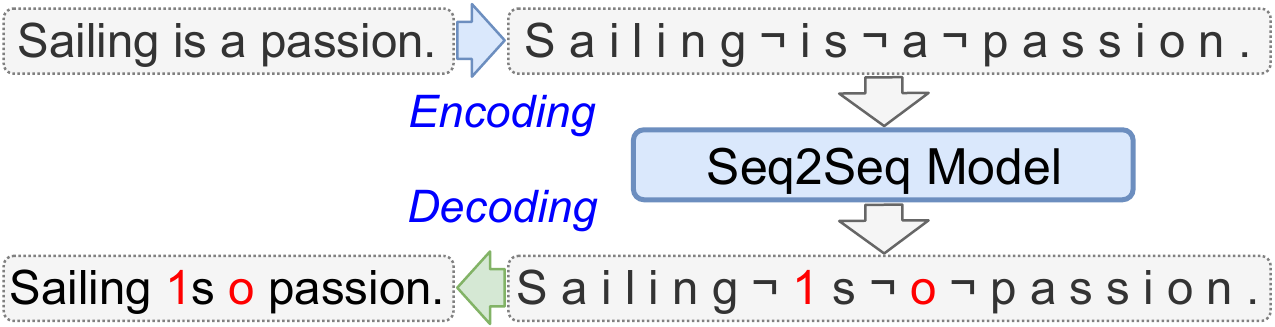}
\caption{
Sentence encoding-decoding schema.
The whitespace characters are first replaced with a placeholder symbol "\unichar{"00AC}".
The sentences are tokenized at the character-level by adding whitespace between every pair of characters.
Decoding reverses this process.
}
\label{fig:onmt_preprocessing}
\end{center}
\end{figure}

To learn our error generators (\S\ref{ssec:noise_generation}), we utilize the OpenNMT\footnote{\url{https://github.com/OpenNMT/OpenNMT-py}} toolkit~\citep{klein-etal-2017-opennmt}.\footnote{We list all non-default hyper-parameters in \Cref{tab:onmt-hyperparams}.}
We encode the input sentence at the character-level before feeding it to the seq2seq model. 
Subsequently, the output produced by the seq2seq model is decoded back to the original form (\Cref{fig:onmt_preprocessing}).

%==================== Parallel Data Extraction ====================
\subsection{Unsupervised Parallel Data Generation}
\label{ssec:parallel_data_generation}

\narrowstyleC
{
Following the approach from \S\ref{ssec:data_generation}, we generated a large parallel corpus $\mathcal{P}$ to train our error generation and correction models.
We sampled 10 million sentences\footnote{Which accounts for about 253 million words.} from the English part of the {\it 1 Billion Word Language Model Benchmark}\footnote{\url{https://www.statmt.org/lm-benchmark}} and used them as the source of error-free text, i.e., the seed corpus $\mathcal{T}$.
We rendered each sentence as an image using the {\it Text Recognition Data Generator} package\footnote{\url{https://pypi.org/project/trdg}}.
We used 90 different fonts for rendering and applied random distortions to the rendered images.
Subsequently, we performed OCR on each image of text using a Python wrapper\footnote{\url{https://github.com/sirfz/tesserocr}} for Tesseract-OCR\footnote{We used Tesseract v4.0 to generate the parallel data set.}~\cite{Smith2007}.
}
We present the distribution of error rates in our noisy corpus in \Cref{fig:error-distrib} (cf. the \emph{digitized text} plot).

%==================== Sequence Labeling Setup ====================
\subsection{Sequence Labeling}
\label{ssec:seq2seq_setup}

\paragraph{Training Setup}
We employed the NAT framework\footnote{\narrowstyleD{\url{https://github.com/mnamysl/nat-acl2020}}} (\Cref{fig:train_arch}) to study the robustness of sequence labeling systems.
Following \citet{akbik-etal-2018-contextual}, we used a combination of \textit{FLAIR} and  \textit{GloVe} embeddings in all experiments.\footnote{Other hyper-parameters also follow~\citet{akbik-etal-2018-contextual}.}
We employed the data augmentation ($\mathcal{L}_\text{AUGM}$) and the stability training ($\mathcal{L}_\text{STAB}$) objectives with default weights ($\alpha=1.0$), as proposed by {\mnt}.
Consistent with prior work, erroneous sentences $\tilde{x}$ were generated dynamically in every epoch.

\paragraph{Tasks}
We experimented with the Named Entity Recognition (NER) and Part-of-Speech Tagging (POST) tasks. 
NER aims to locate all named entity mentions in text and classify them into predefined classes, e.g., person names, locations, and organizations.
POST is the process of tagging each word in the text with the corresponding part of speech. 

\paragraph{Evaluation Setup}

The evaluation pipeline is shown in \Cref{fig:eval_arch}.
Following \citet{akbik-etal-2018-contextual}, we report the entity-level micro-average F1 score for NER and the accuracy for POST.
{
\begin{figure}[ht]
\begin{center}
\includegraphics[width=1.0\columnwidth]{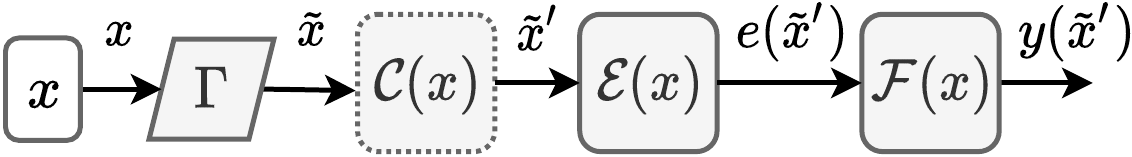}
\caption{
Evaluation pipeline.
$\Gamma$ is a noising process that transforms $x$ into $\tilde{x}$.
$\mathcal{C}(x)$ is an optional text correction module that returns $\tilde{x}'$ ($\tilde{x}'\myeq\tilde{x}$, if $\mathcal{C}(x)$ is absent).
$\mathcal{E}(x)$ is an embedding matrix.
$\mathcal{F}(x)$ is a sequence labeling model.
$e(\tilde{x}')$ and $y(\tilde{x}')$ are the embeddings and the output of the model for $\tilde{x}'$, respectively. 
}
\label{fig:eval_arch}
\end{center}
\end{figure}
}
%

%==================== Baselines ====================
\subsection{Baselines}
\label{ssec:baselines}

\paragraph{Error Generation}

We compared our error generator with the OCR-aware noise model from {\mnt}.
We used the noisy part of the parallel corpus $\mathcal{P}$ to estimate the confusion matrix employed by this baseline.
Moreover, in the NLM experiment (\S\ref{ssec:eval-noisy-lm}), we also evaluated the vanilla error model proposed by {\mnt}.

\paragraph{Error Correction}

To evaluate error correction, we trained the sequence labeling models using the standard objective ($\mathcal{L}_{0}$) and employed the text correction method on the erroneous input before feeding it to the network (\Cref{fig:eval_arch}).

We examined \textit{Natas}\footnote{\url{https://github.com/mikahama/natas}}, the seq2seq OCR post-correction method proposed by \natast. 
We trained context-free error correction models compatible with Natas using our parallel corpus (\S\ref{ssec:parallel_data_generation}).
Moreover, we also employed the widely adopted spell checker \textit{Hunspell}\footnote{\url{https://hunspell.github.io}}.

%==================== Data Sets ====================
\subsection{Data Sets}
\label{ssec:noisy_datasets}

\paragraph{Original Benchmarks}
For NER, we employed the CoNLL 2003 data set~\citep{tjong-kim-sang-de-meulder-2003-introduction}.
To evaluate POST, we utilized the Universal Dependency Treebank (UD English EWT;~\citealp{silveira-etal-2014-gold}).
We present the detailed statistics of both data sets in \Cref{tab:data_sets}.

\paragraph{Noisy Benchmarks}

{\renewcommand{\arraystretch}{1.0}\setlength{\tabcolsep}{0.0pt}
\begin{table}[!ht]
\centering\small
\begin{tabular}{@{}L{0.11}C{0.08}C{0.09}C{0.1}C{0.1}@{}}
\toprule
Data Set & Geom. Distort. & Pixel-level Noise & {CoNLL 2003} & {UD English EWT}\\
\midrule
Tesseract 3$^\clubsuit$    & $\xmark$ & $\xmark$ & 22.72\% & 23.31\% \\
Tesseract 4$^\diamondsuit$ & $\cmark$ & $\cmark$ & 16.35\% & 22.12\% \\
Tesseract 4$^\heartsuit$   & $\cmark$ & $\xmark$ & 14.89\% & 20.38\% \\
Tesseract 4$^\spadesuit$   & $\xmark$ & $\cmark$ & 3.53\%  & 5.83\% \\
Typos & \na & \na & 15.53\% & 15.22\%\\
\bottomrule
\end{tabular}
\caption{
\narrowstyleB
{
The noisy sequence labeling data sets that we generated either by applying OCR on rendered sentences from an original benchmark (first four rows) or by inducing misspellings (last row).
We generated multiple variants of the former data sets by combining geometrical distortions and pixel-level noise induction.
The last two columns present the token error rates (the column headers indicate the names of the original benchmarks).
}}
\label{tab:ocr_data_sets}
\end{table}}

Unfortunately, we did not find any publicly available noisy sequence labeling data set that could be used to benchmark different methods for improving robustness.
To this end, we generated several noisy versions of the original sequence labeling data sets (\Cref{tab:ocr_data_sets}).
We extracted the sentences from each original benchmark and applied the procedure described in \S\ref{ssec:data_generation}.\footnote{We directly applied both Tesseract v3.04 and v4.0. We used different sets of distortions and image backgrounds than those employed to generate parallel training data.}
We transferred the word-level annotations as described in \S\ref{ssec:data_alignment}.
Finally, we produced the data in the CoNLL format (\Cref{tab:conll-format}).

Moreover, to evaluate the transferability of error generators, we followed {\mnt} and synthetically induced misspellings to the error-free data sets.
To this end, we used the lookup tables of possible lexical replacements released by \citet{DBLP:conf/iclr/BelinkovB18} and~\citet{piktus-etal-2019-misspelling}.\footnote{We merged both sets of misspellings for evaluation.}

%--------------------------------------------------------------
%==================== Experimental Results ====================
%--------------------------------------------------------------

\section{Experimental Results}
\label{sec:experimental_results}

%==================== Comparison of Different Noise Generation Approaches ====================

\subsection{Empirical Noise Generation Approaches}
\label{ssec:eval-err-gen}

{\renewcommand{\arraystretch}{0.95}\setlength{\tabcolsep}{3.5pt}
\newcommand{\ColWidthA}{0.065}
\begin{table*}[htbp]
%
% ---- English CoNLL 2003 ----
%
\begin{subtable}{\textwidth}\centering\small
\begin{tabular}{@{}C{\ColWidthA}L{0.24}L{0.08}*{5}{C{0.1}}@{}}
\toprule
Training Loss & Noise Model & Correction Method & Original Data & Tesseract 3$^\clubsuit$ & Tesseract 4$^\diamondsuit$ & Tesseract 4$^\heartsuit$ & Tesseract 4$^\spadesuit$ \\ 
\midrule
\multirow{3}{\ColWidthA\textwidth}[0pt]{$\mathcal{L}_0$} & \na & \none & \insep{\meanstddev{92.54}{0.08}} & \meanstddev{80.48}{0.09} & \meanstddev{84.71}{0.19} & \meanstddev{85.62}{0.08} & \meanstddev{91.50}{0.08} \\
& \na & {Hunspell} & \insep{\meanstddev{92.54}{0.08}} & \best{\meanstddev{82.17}{0.11}} & \best{\meanstddev{85.80}{0.11}} & \best{\meanstddev{86.70}{0.07}} & \best{\meanstddev{91.73}{0.11}} \\
& \na & {Natas} & \insep{\meanstddev{92.54}{0.08}} & \meanstddev{77.80}{0.19} & \meanstddev{84.50}{0.11} & \meanstddev{85.24}{0.10} & \meanstddev{91.33}{0.13} \\ 
\midrule
\multirow{3}{\ColWidthA\textwidth}[0pt]{$\mathcal{L}_\text{AUGM}$} 
& confusion matrix (\S\ref{ssec:cmx-model}) & \none & \meanstddev{92.56}{0.06} & \insep{\meanstddev{85.29}{0.16}} & \meanstddev{88.62}{0.08} & \meanstddev{89.19}{0.12} & \meanstddev{92.04}{0.07} \\
& \tokseq & \none & \insep{\meanstddev{92.76}{0.07}} & \best{\meanstddev{85.38}{0.16}} & \best{\meanstddev{89.39}{0.17}} & \best{\meanstddev{89.99}{0.22}} & \insep{\meanstddev{92.37}{0.10}} \\ 
& \charseq & \none & \best{\meanstddev{92.81}{0.11}} & \meanstddev{84.38}{0.15} & \meanstddev{88.96}{0.18} & \insep{\meanstddev{89.67}{0.26}} & \best{\meanstddev{92.44}{0.17}} \\ 
\midrule
\multirow{3}{\ColWidthA\textwidth}[0pt]{$\mathcal{L}_\text{STAB}$} 
& confusion matrix (\S\ref{ssec:cmx-model}) & \none & \meanstddev{92.23}{0.12} & \best{\meanstddev{84.49}{0.10}} & \meanstddev{87.58}{0.13} & \meanstddev{88.40}{0.20} & \meanstddev{91.65}{0.14} \\ 
& \tokseq & \none & \insep{\meanstddev{92.24}{0.18}} & \insep{\meanstddev{84.25}{0.23}} & \best{\meanstddev{88.24}{0.25}} & \best{\meanstddev{88.91}{0.21}} & \insep{\meanstddev{91.86}{0.16}} \\ 
& \charseq & \none & \best{\meanstddev{92.45}{0.12}} & \meanstddev{83.89}{0.30} & \insep{\meanstddev{88.14}{0.23}} & \insep{\meanstddev{88.88}{0.11}} & \best{\meanstddev{91.99}{0.11}} \\
\bottomrule
\end{tabular}
\caption{English CoNLL 2003}
\end{subtable}
\par\smallskip % force a bit of vertical whitespace
%
% ---- UD English EWT ----
%
\begin{subtable}{\textwidth}\centering\small
\begin{tabular}{@{}C{\ColWidthA}L{0.24}L{0.08}*{5}{C{0.1}}@{}}
\toprule
\multirow{3}{\ColWidthA\textwidth}[0pt]{$\mathcal{L}_0$} & \na & \none & \insep{\meanstddev{96.96}{0.04}} & \meanstddev{86.75}{0.16} & \meanstddev{86.97}{0.14} & \meanstddev{88.30}{0.16} & \meanstddev{94.34}{0.07} \\ 
& \na & {Hunspell} & \insep{\meanstddev{96.96}{0.04}} & \meanstddev{87.53}{0.14} & \meanstddev{86.74}{0.14} & \meanstddev{88.12}{0.16} & \meanstddev{94.49}{0.08} \\
& \na & {Natas} & \insep{\meanstddev{96.96}{0.04}} & \best{\meanstddev{88.98}{0.10}} & \best{\meanstddev{88.94}{0.14}} & \best{\meanstddev{89.68}{0.16}} & \best{\meanstddev{95.11}{0.08}} \\ 
\midrule
\multirow{3}{\ColWidthA\textwidth}[0pt]{$\mathcal{L}_\text{AUGM}$} 
& confusion matrix (\S\ref{ssec:cmx-model}) & \none & \best{\meanstddev{96.90}{0.06}} & \insep{\meanstddev{91.35}{0.13}} & \meanstddev{92.12}{0.14} & \meanstddev{92.99}{0.21} & \meanstddev{96.17}{0.07} \\ 
& \tokseq & \none & \meanstddev{96.76}{0.04} & \best{\meanstddev{91.44}{0.11}} & \best{\meanstddev{93.65}{0.13}} & \best{\meanstddev{94.19}{0.10}} & \insep{\meanstddev{96.26}{0.07}} \\ 
& \charseq & \none & \meanstddev{96.78}{0.06} & \meanstddev{90.92}{0.08} & \meanstddev{93.37}{0.08} & \insep{\meanstddev{94.10}{0.03}} & \best{\meanstddev{96.27}{0.03}} \\ 
\midrule
\multirow{3}{\ColWidthA\textwidth}[0pt]{$\mathcal{L}_\text{STAB}$} 
& confusion matrix (\S\ref{ssec:cmx-model}) & \none & \best{\meanstddev{96.80}{0.04}} & \meanstddev{91.16}{0.07} & \meanstddev{91.93}{0.11} & \meanstddev{92.77}{0.10} & \meanstddev{96.06}{0.02} \\
& \tokseq & \none & \meanstddev{96.65}{0.07} & \best{\meanstddev{91.36}{0.12}} & \best{\meanstddev{93.34}{0.09}} & \best{\meanstddev{93.97}{0.05}} & \insep{\meanstddev{96.14}{0.07}} \\ 
& \charseq & \none & \meanstddev{96.67}{0.05} & \meanstddev{90.70}{0.14} & \meanstddev{93.05}{0.17} & \meanstddev{93.71}{0.13} & \best{\meanstddev{96.15}{0.05}} \\ 
\bottomrule
\end{tabular}
\caption{UD English EWT}
\end{subtable}

\caption{
\narrowstyleD
{
Comparison of error generation (\S\ref{ssec:eval-err-gen}) and error correction (\S\ref{ssec:eval-err-corr}) approaches on the original and noisy English CoNLL 2003 and the UD English EWT test sets (\S\ref{ssec:noisy_datasets}).
We report mean and standard deviation F1 scores (CoNLL 2003) and accuracies (UD English EWT) over five runs with different random initialization.
$\mathcal{L}_0$, $\mathcal{L}_\text{AUGM}$, $\mathcal{L}_\text{STAB}$ is the standard, the data augmentation, and the stability objective, respectively~\mnp.
\textbf{Bold} values indicate top results (within the models trained using the same objective) that are statistically inseparable (Welch's t-test; $p<0.05$).
}}
\label{tab:eval-err-gen-corr}
\end{table*}}

In this experiment, we compared the NAT models that employed either our seq2seq noise generators or the baseline error models (\Cref{tab:eval-err-gen-corr}).
In this evaluation scenario, we do not employ $\mathcal{C}(x)$ (\Cref{fig:eval_arch}).

Our error generators outperformed the OCR-aware confusion matrix-based model on the noisy benchmarks generated using the Tesseract 4 engine.
The advantage of our method was less emphasized in the case of the \emph{Tesseract 3$^\clubsuit$} data sets.
The {\tok} translation method performed better than the {\ch} variant, while the latter was more efficient when the error rate of the input was lower (cf. the \emph{original data} and the \emph{Tesseract 4$^\spadesuit$} columns), although it often struggled with translating long sentences.
Moreover, data augmentation generally outperformed stability training, which is consistent with the observation from~{\mnt}.

Furthermore, we observe a slight decrease in accuracy on the original UD English EWT with both auxiliary objectives. 
We believe that this was caused by the different proportions of the tokens that were perturbed during training by our seq2seq error generators (e.g., 18\% and 19.5\% in the case of our {\tok} model for CoNLL2003 and UD English EWT, respectively).
The trade-off between accuracy for clean and noisy data has thus been shifted towards the latter. 
We also notice a greater advantage of the seq2seq method over the baseline on the noisy UD English EWT data sets.

Additionally, in \S\ref{sec:eval-low-res}, we analyze the relationship between the size of the parallel corpus used for training and the F1 score of the NER task. 

%==================== Error Generation vs. Error Correction ====================
\subsection{Error Generation vs. Error Correction}
\label{ssec:eval-err-corr}

We compared the NAT approach with the baseline correction methods (\S\ref{ssec:baselines}).
Preliminary experiments revealed that these baselines underperformed due to the \emph{overcorrection} problem.
To make them more competitive, we extended their default dictionaries by adding all tokens from the corresponding test sets for evaluation.
Although the vocabulary of a test set could rarely be entirely determined, this setting would simulate a scenario where accurate in-domain vocabularies could be exploited.

\Cref{tab:eval-err-gen-corr} includes the results of this experiment.
As expected, although more general, error correction techniques were outperformed by the NAT approach regardless of the noising method used.
Surprisingly, Hunspell performed better than Natas on CoNLL 2003.
We carried out a thorough inspection of the results of both methods and found out that Natas, although generally more accurate, had problems with recognizing tokens that were a part of entities.
This behavior could be a flaw of data-driven error correction methods, as the entities are relatively rare in written text and are often out-of-vocabulary tokens \citep{Alex:2014:ERQ:2595188.2595214}.

%==================== Noisy Language Modeling ====================
\subsection{Noisy Language Modeling}
\label{ssec:eval-noisy-lm}

{\renewcommand{\arraystretch}{1.0}\setlength{\tabcolsep}{6.0pt}
\newcommand{\ColWidthA}{0.065}
\newcommand{\ColWidthB}{0.19}
\begin{table*}[htb]
\centering \small 
\begin{tabular}{@{}C{\ColWidthA}L{\ColWidthB}C{0.04}*{5}{C{0.1}}@{}}
\toprule
% ---- English CoNLL 2003 ----
Training Loss & Noise Model & NLM & Original Data & Tesseract 3$^\clubsuit$ & Tesseract 4$^\diamondsuit$ & Tesseract 4$^\heartsuit$ & Tesseract 4$^\spadesuit$ \\
\midrule
\multirow{2}{\ColWidthA\textwidth}[0pt]{$\mathcal{L}_0$} & \na & $\xmark$ & \best{\meanstddev{92.54}{0.08}} & \meanstddev{80.48}{0.09} & \meanstddev{84.71}{0.19} & \meanstddev{85.62}{0.08} & \meanstddev{91.50}{0.08} \\
& \na & $\cmark$ & \meanstddev{92.09}{0.07} & \best{\meanstddev{83.83}{0.21}} & \best{\meanstddev{88.17}{0.12}} & \best{\meanstddev{88.71}{0.17}} & \best{\meanstddev{91.68}{0.09}} \\
\midrule
\multirow{3}{\ColWidthA\textwidth}[0pt]{$\mathcal{L}_\text{AUGM}$} 
& confusion matrix (\S\ref{ssec:cmx-model}) & $\xmark$ & \best{\meanstddev{92.56}{0.06}} & \meanstddev{85.29}{0.16} & \meanstddev{88.62}{0.08} & \meanstddev{89.19}{0.12} & \meanstddev{92.04}{0.07} \\
& vanilla (\S\ref{ssec:cmx-model}) & $\xmark$ & \meanstddev{92.39}{0.11} & \meanstddev{85.59}{0.23} & \meanstddev{88.01}{0.17} & \meanstddev{88.65}{0.20} & \meanstddev{91.93}{0.13} \\
& vanilla (\S\ref{ssec:cmx-model}) & $\cmark$ & \meanstddev{92.45}{0.05} & \best{\meanstddev{87.28}{0.19}} & \best{\meanstddev{90.12}{0.19}} & \best{\meanstddev{90.43}{0.19}} & \best{\meanstddev{92.17}{0.05}} \\
\midrule
\multirow{3}{\ColWidthA\textwidth}[0pt]{$\mathcal{L}_\text{STAB}$} 
& confusion matrix (\S\ref{ssec:cmx-model}) & $\xmark$ & \best{\meanstddev{92.23}{0.12}} & \meanstddev{84.49}{0.10} & \meanstddev{87.58}{0.13} & \meanstddev{88.40}{0.20} & \best{\meanstddev{91.65}{0.14}} \\
%%%
& vanilla (\S\ref{ssec:cmx-model}) & $\xmark$ & \meanstddev{92.04}{0.06} & \meanstddev{84.63}{0.17} & \meanstddev{87.24}{0.24} & \meanstddev{88.02}{0.10} & \insep{\meanstddev{91.52}{0.12}} \\
& vanilla (\S\ref{ssec:cmx-model}) & $\cmark$ & \meanstddev{91.85}{0.07} & \best{\meanstddev{86.79}{0.11}} & \best{\meanstddev{89.32}{0.12}} & \best{\meanstddev{89.77}{0.05}} & \insep{\meanstddev{91.51}{0.07}} \\
\bottomrule
\end{tabular}
\caption{
\narrowstyleC
{
Comparison of the NAT approach with and without our NLM embeddings (\S\ref{ssec:eval-noisy-lm}) on the English CoNLL 2003 test set (\S\ref{ssec:noisy_datasets}).
We report mean and standard deviation F1 scores over five runs with different random initialization.
$\mathcal{L}_0$, $\mathcal{L}_\text{AUGM}$, $\mathcal{L}_\text{STAB}$ is the standard, the data augmentation, and the stability objective, respectively \mnp.
The \emph{NLM} column indicates whether the model employed our NLM embeddings.
\textbf{Bold} values indicate top results (within the models trained using the same objective) that are statistically inseparable (Welch's t-test; $p<0.05$).
}}
\label{tab:eval-noisy-lm}
\end{table*}}

FLAIR~\cite{akbik-etal-2018-contextual} learns a bidirectional LM to represent sequences of characters.
We used the target side of our parallel data corpus (\S\ref{ssec:parallel_data_generation}) to re-train FLAIR embeddings on the noisy digitized text.\footnote{The hyper-parameters were consistent with prior work.}
Subsequently, we compared the accuracy of the vanilla NAT models (\S\ref{ssec:cmx-model}) that employed either the pre-trained or our NLM embeddings.
Moreover, we do not use $\mathcal{C}(x)$ in this scenario (\Cref{fig:eval_arch}).

Note that the noise model and the embeddings are two distinct components of the NAT architecture ($\Gamma$ and $\mathcal{E}(x)$ in \Cref{fig:train_arch}, respectively) and therefore they could be easily combined.
However, in this work, we do not mix our NLM with empirically estimated error models to avoid the twofold empirical error modeling effect. We leave the evaluation of this combination to future work.

\Cref{tab:eval-noisy-lm} summarizes the results of this experiment.
Our method significantly improved the accuracy across all training objectives, even when we employed exclusively the standard training objective for the sequence labeling task ($\mathcal{L}_0$).
Surprisingly, we also achieved evident improvements for the noisy data set generated using the Tesseract~3 engine, which confirms that NLM embeddings can model the features of erroneous tokens even in the out-of-domain scenarios.
On the other hand, the NLM slightly decreased the accuracy on the original data for the standard training objective.
We plan to investigate this effect in future work by eliminating possible differences in the pre-training procedure and comparing our NLM against a model trained on the original error-free text corpus instead of using the embeddings from \citet{akbik-etal-2018-contextual}.
%

%==================== Human-generated Errors ====================
\subsection{Human-Generated Errors}
\label{ssec:eval-typos}

In this experiment, we evaluated the utility of our seq2seq error generators learned to model OCR noise (\S\ref{ssec:eval-err-gen}) and our NLM embeddings (\S\ref{ssec:eval-noisy-lm}) in a scenario where the input contains human-generated errors.
For evaluation, we used the noisy data sets with synthetically induced misspellings (\S\ref{ssec:noisy_datasets}).
We do not employ $\mathcal{C}(x)$ in this scenario (\Cref{fig:eval_arch}).

{\renewcommand{\arraystretch}{1.0}\setlength{\tabcolsep}{1pt}
\newcommand{\ColWidthA}{0.16}
\newcommand{\ColWidthB}{0.13}
\begin{table*}[!ht]\centering
%
% ---- NLM Embeddings ----
%
\begin{subtable}[t]{0.444\textwidth}\centering\small
\begin{tabular}{@{}C{\ColWidthA}L{0.42}C{0.1}*{1}{C{0.25}}@{}}
\toprule
Training Loss & Noise Model & NLM & English CoNLL 2003 \\
\midrule
\multirow{2}{\ColWidthA\textwidth}[0pt]{$\mathcal{L}_0$} & \na & $\xmark$ & \meanstddev{88.79}{0.07} \\
& \na & $\cmark$ & \best{\meanstddev{89.60}{0.24}} \\
\midrule
\multirow{3}{\ColWidthA\textwidth}[0pt]{$\mathcal{L}_\text{AUGM}$} 
& confusion matrix (\S\ref{ssec:cmx-model}) & $\xmark$ & \meanstddev{90.82}{0.12}\\
& vanilla (\S\ref{ssec:cmx-model}) & $\xmark$ & \meanstddev{90.77}{0.14} \\
& vanilla (\S\ref{ssec:cmx-model}) & $\cmark$ & \best{\meanstddev{91.10}{0.05}} \\
\midrule
\multirow{3}{\ColWidthA\textwidth}[0pt]{$\mathcal{L}_\text{STAB}$} 
& confusion matrix (\S\ref{ssec:cmx-model}) & $\xmark$ & \meanstddev{90.30}{0.13} \\
& vanilla (\S\ref{ssec:cmx-model}) & $\xmark$ & \meanstddev{90.34}{0.06} \\
& vanilla (\S\ref{ssec:cmx-model}) & $\cmark$ & \best{\meanstddev{90.53}{0.07}} \\
\bottomrule
\end{tabular}
\caption{NLM Embeddings}
\label{tab:eval-typos-nlm}
\end{subtable}
%
% ---- Empirical Error Generators ----
%
%\hfil
\begin{subtable}[t]{0.55\textwidth}\centering\small
\begin{tabular}{@{}C{\ColWidthB}L{0.43}*{2}{C{0.19}}@{}}
\toprule
Training Loss & Noise Model & English CoNLL 2003 & UD English EWT \\
\midrule
\multirow{1}{\ColWidthB\textwidth}[0pt]{$\mathcal{L}_0$} & \na & {\meanstddev{88.79}{0.07}} & \meanstddev{90.54}{0.11}\\
\midrule
\multirow{3}{\ColWidthB\textwidth}[0pt]{$\mathcal{L}_\text{AUGM}$} 
& confusion matrix (\S\ref{ssec:cmx-model}) & \insep{\meanstddev{90.82}{0.12}} & \best{\meanstddev{93.63}{0.11}} \\
& \tokseq & \best{\meanstddev{90.92}{0.13}} & \meanstddev{92.87}{0.08} \\
& \charseq & \insep{\meanstddev{90.77}{0.19}} & \meanstddev{92.68}{0.09} \\
\midrule
\multirow{3}{\ColWidthB\textwidth}[0pt]{$\mathcal{L}_\text{STAB}$} 
& confusion matrix (\S\ref{ssec:cmx-model}) & \best{\meanstddev{90.30}{0.13}} & \best{\meanstddev{93.37}{0.05}} \\
& \tokseq & \insep{\meanstddev{90.19}{0.12}} & \meanstddev{92.79}{0.08} \\
& \charseq & \insep{\meanstddev{90.15}{0.16}} & \meanstddev{92.42}{0.11} \\
\bottomrule
\end{tabular}
\caption{Empirical Error Generation Methods}
\label{tab:eval-typos-errgen}
\end{subtable}
\caption{
\narrowstyleB
{
Transferability of the methods learned to model OCR noise to the distribution of the human-generated errors (\S\ref{ssec:eval-typos}):
\begin{enumerate*}[label=\textbf{(\alph*)}]
\item Comparison of the NAT approach with and without our NLM embeddings on the English CoNLL 2003 test set with human-generated errors.
\item Comparison of empirical error generation approaches on the English CoNLL 2003 and the UD English EWT test sets with human-generated errors.
\end{enumerate*}
We report mean and standard deviation F1 scores (CoNLL 2003) and accuracies (UD English EWT) over five runs with different random initialization.
$\mathcal{L}_0$, $\mathcal{L}_\text{AUGM}$, $\mathcal{L}_\text{STAB}$ is the standard, the data augmentation, and the stability objective, respectively~\mnp.
The \emph{NLM} column indicates whether the model employed our NLM embeddings.
\textbf{Bold} values indicate top results (within the models trained using the same objective) that are statistically inseparable (Welch's t-test; $p<0.05$).
}
}
\label{tab:eval-typos}
\end{table*}}

\Cref{tab:eval-typos} summarizes the results of this experiment.
The models with our NLM embeddings outperformed the baselines for all training objectives (\Cref{tab:eval-typos-nlm}).
The seq2seq error generation approach performed on par with the confusion matrix-based models on the CoNLL 2003 data set, while the latter achieved better accuracy on the UD English EWT data set (\Cref{tab:eval-typos-errgen}).

We believe that this difference was caused by the discrepancy between the data distributions.
Note that although the data used in this experiment reflects the patterns of human-generated errors, the distribution of these errors does not necessarily follow the natural distribution of human-generated errors, as it was synthetically generated using a fixed replacement probability that was uniform across all candidates.\footnote{For comparison, we visualized the error distributions of our noisy benchmarks in \Cref{fig:error-distrib2}.}
Nevertheless, our methods proved to be beneficial in this scenario, which would suggest that the errors made by human writers and by the text recognition engines have common characteristics that were exploited by our method.

\onlyinsubfile{
\bibliographystyle{../acl_natbib}
\bibliography{../anthology,../acl2021}
}

%% file: sections/conclusions.tex
\makeatletter
\let\maintitle\@title
\title{
\maintitle\\
(Conclusions)
}
\makeatother

\maketitle

\section{Conclusions}
\label{sec:conclusions}
In this work, we studied the task of performing sequence labeling on noisy digitized and human-generated text.
We extended the NAT approach and proposed the empirical error generator that performs the translation from error-free to erroneous text (\S\ref{ssec:noise_generation}).
To train our generator, we developed an unsupervised parallel data synthesis method (\S\ref{ssec:data_generation}).
Analogously, we produced several realistic noisy evaluation benchmarks (\S\ref{ssec:noisy_datasets}).
Moreover, we introduced the NLM embeddings (\S\ref{ssec:noisy_lm}) that overcome the data sparsity problem of natural language.

Our approach outperformed the baseline noise induction and error correction methods, improving the accuracy of the noisy neural sequence labeling task (\S\ref{ssec:eval-err-gen}-\ref{ssec:eval-noisy-lm}).
Furthermore, we demonstrated that our methods are transferable to the out-of-domain scenarios - human-generated errors (\S\ref{ssec:eval-typos}) and the noise induced by a different OCR engine (\S\ref{ssec:eval-err-gen}, \ref{ssec:eval-noisy-lm}).
We incorporated our approach into the NAT framework and make the code, embeddings, and scripts from our experiments publicly available.

\citet{grundkiewicz-junczys-dowmunt-2019-minimally} showed that that unsupervised systems benefit from domain adaptation on authentic labeled data. 
For future work, we plan to fine-tune NAT models pre-trained on synthetic samples using the labeled data generated directly by the noising process.

%\onlyinsubfile{
%\bibliographystyle{../acl_natbib}
%\bibliography{../anthology,../acl2021}
%}

%% file: sections/appendix.tex
\makeatletter
\let\maintitle\@title
\title{
\maintitle~(Appendix)
}
\makeatother

\maketitle

%==================== Relationship with the Size of the Parallel Corpus ====================
\section{Relationship with the Corpus Size}
\label{sec:eval-low-res}

Empirical error generators are especially beneficial when we can approximate the noise distribution to be encountered at test time.
In this experiment, we aimed to answer the question, how much parallel training data is required to train a solid seq2seq error generation model.

\Cref{fig:eval-low-res} shows that the NAT models that used our seq2seq error generator performed better than those employing the baseline vanilla error model proposed by {\mnt} for all noisy benchmarks that were generated using the \emph{Tesseract 4} OCR engine.
The improvements were observed even when we used as few as 1000 parallel training sentences.
Our method also outperformed the baseline on the original CoNLL 2003 benchmark.
On the contrary, the accuracy of models trained using our generator fell slightly behind the baseline on the \emph{Tesseract 3$^\clubsuit$} and \emph{Typos} data sets.
{\newcommand{\FigWidth}{0.5\columnwidth}
\begin{figure*}[htb]
\begin{center}
\begin{subfigure}[!htbp]{\FigWidth}
\includegraphics[width=1.0\columnwidth]{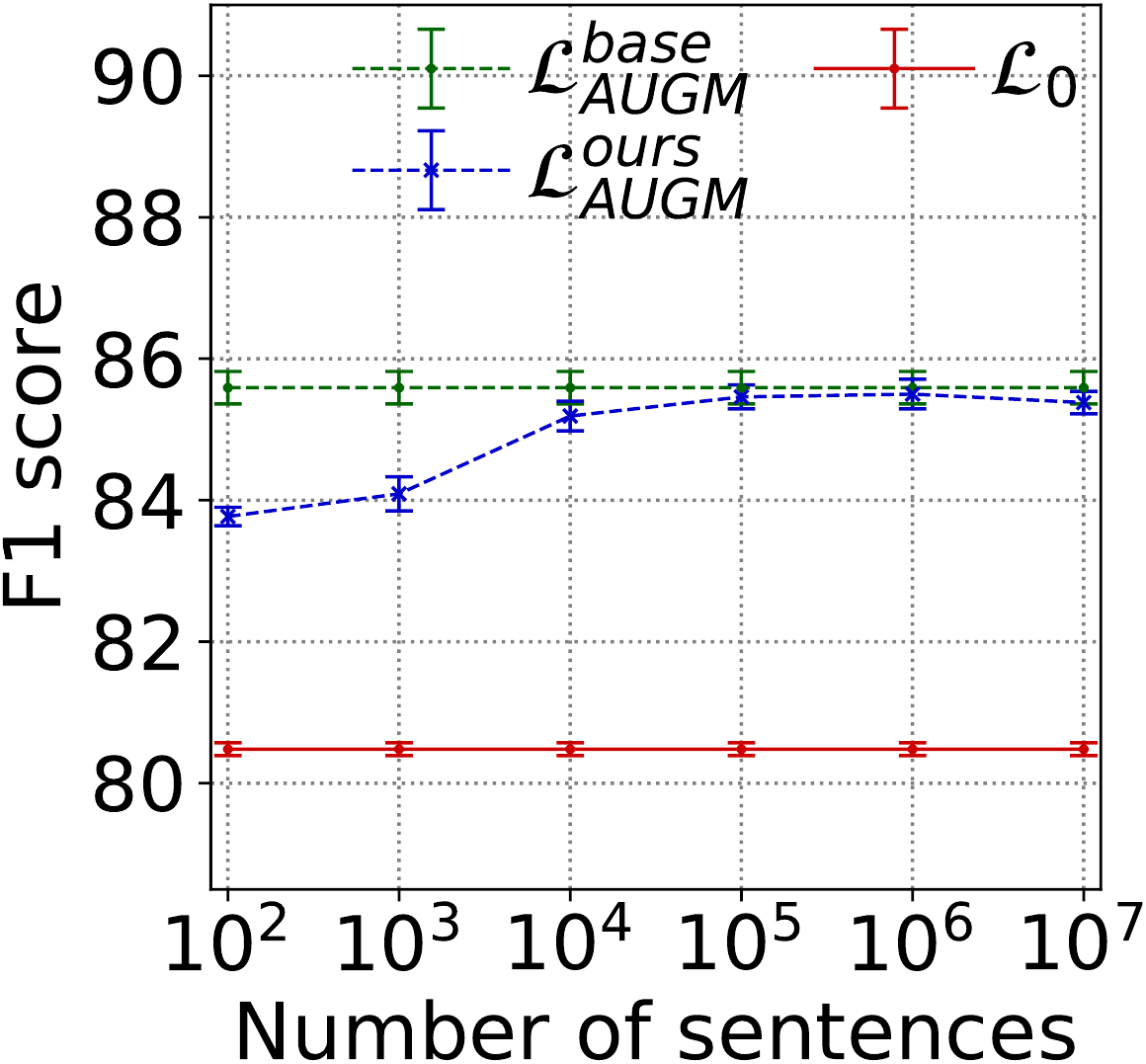}
\caption{$\mathcal{L}_\text{AUGM}$ (Tesseract 3$^\clubsuit$).}
\end{subfigure}
\begin{subfigure}[!htbp]{\FigWidth}
\includegraphics[width=1.0\columnwidth]{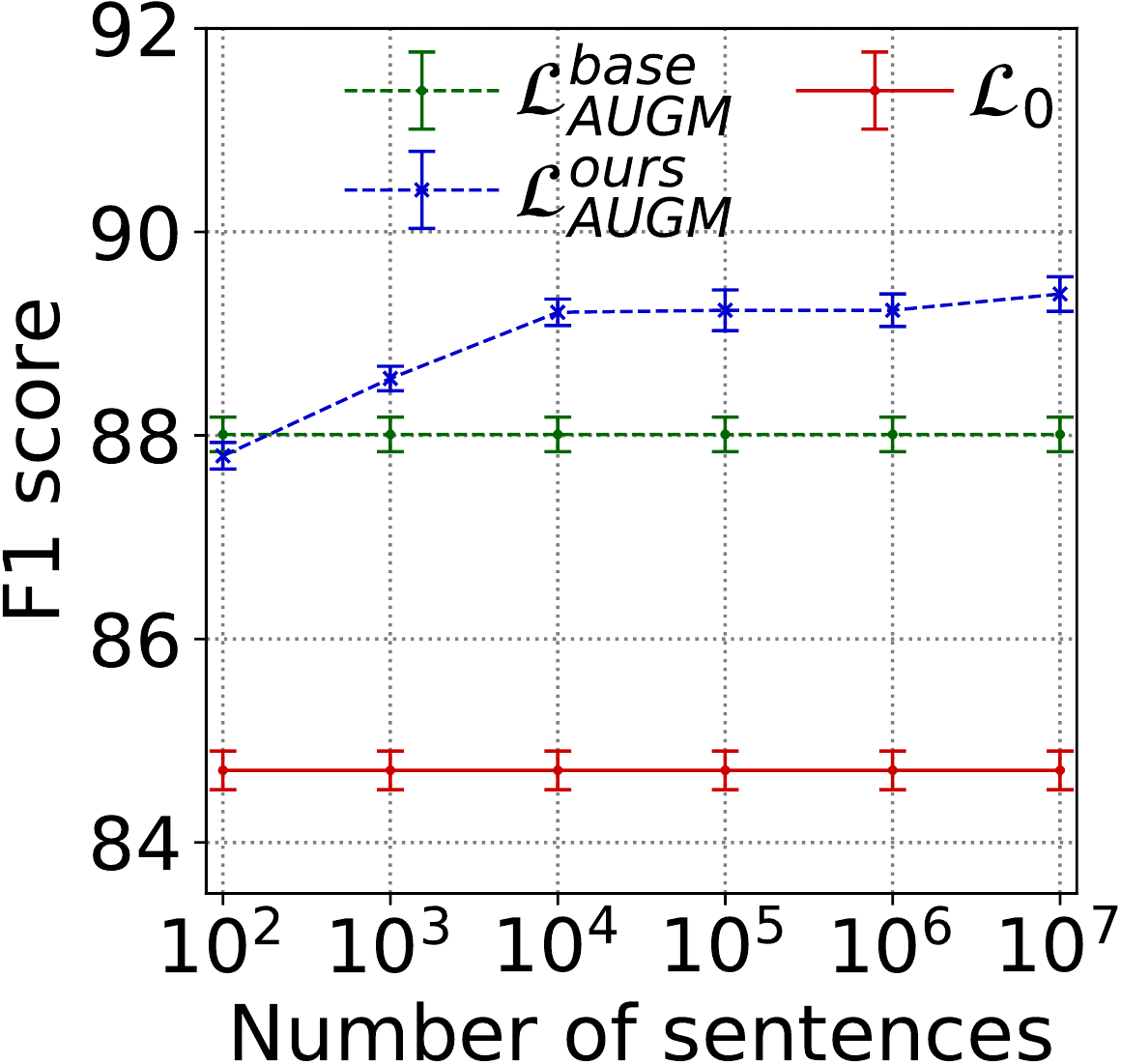}
\caption{$\mathcal{L}_\text{AUGM}$ (Tesseract 4$^\diamondsuit$).}
\end{subfigure}
\begin{subfigure}[!htbp]{\FigWidth}
\includegraphics[width=1.0\columnwidth]{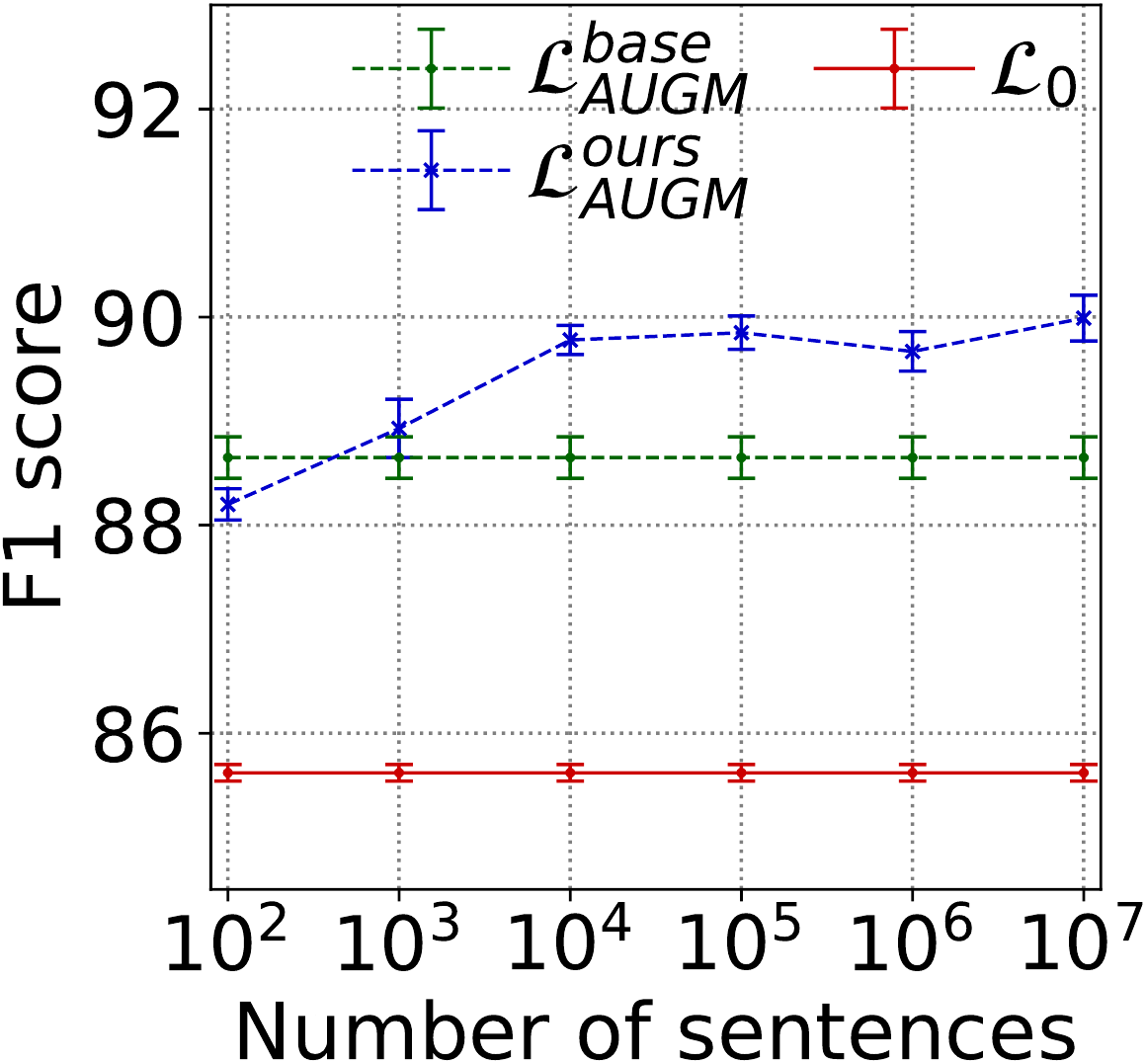}
\caption{$\mathcal{L}_\text{AUGM}$ (Tesseract 4$^\heartsuit$).}
\end{subfigure}
\begin{subfigure}[!htbp]{\FigWidth}
\includegraphics[width=1.0\columnwidth]{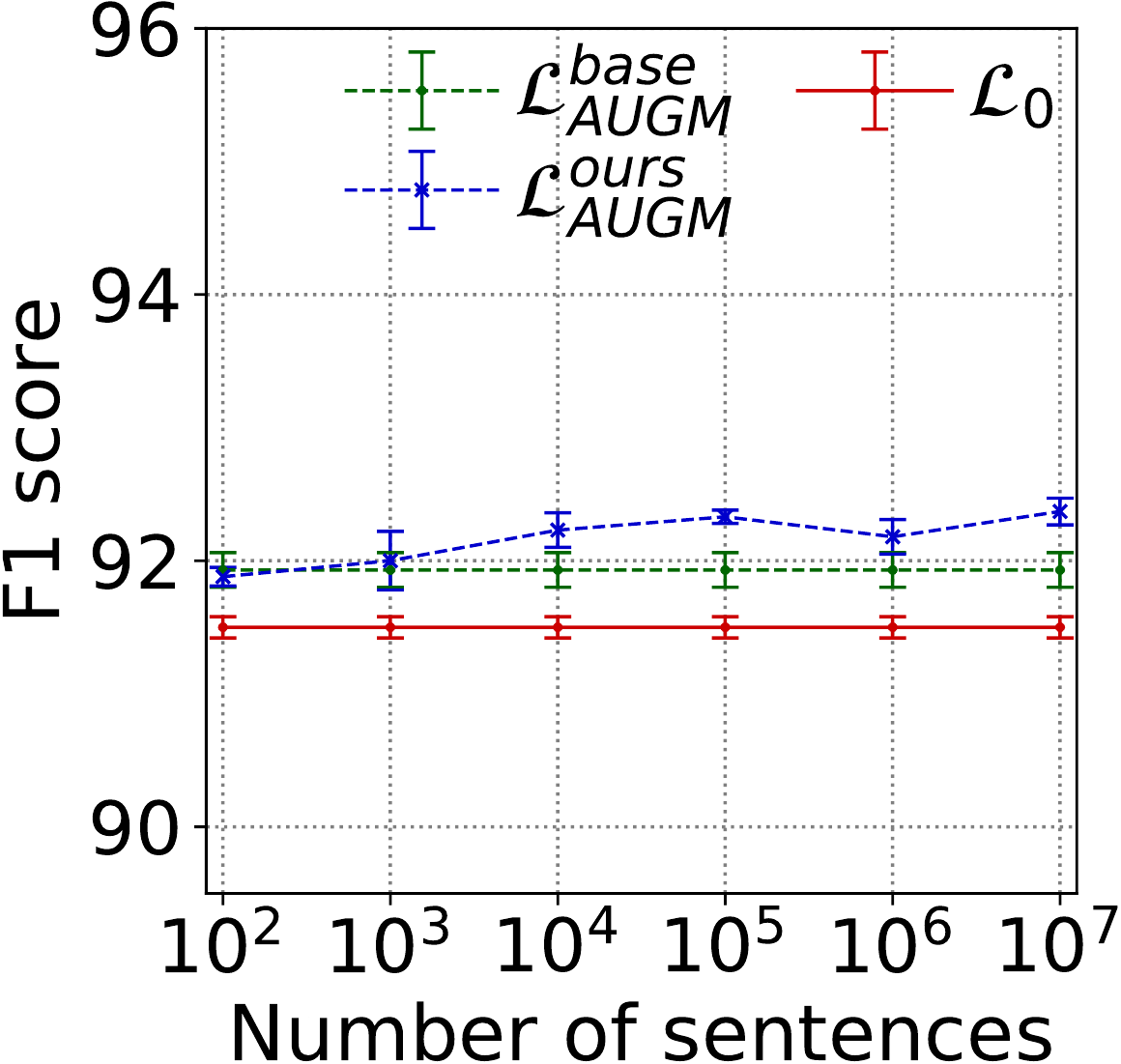}
\caption{$\mathcal{L}_\text{AUGM}$ (Tesseract 4$^\spadesuit$).}
\end{subfigure}
\par\smallskip % force a bit of vertical whitespace
\begin{subfigure}[!htbp]{\FigWidth}
\includegraphics[width=1.0\columnwidth]{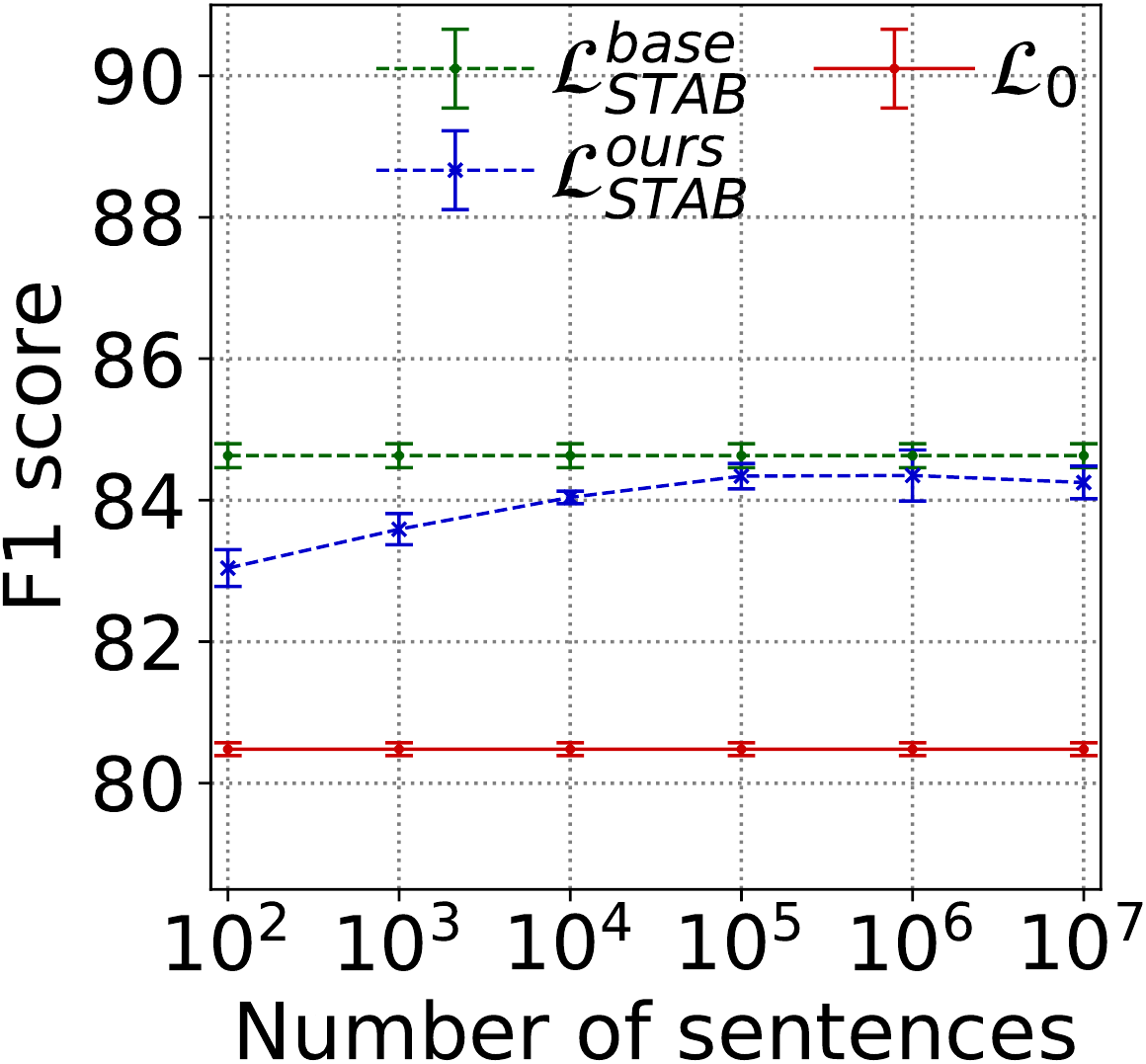}
\caption{$\mathcal{L}_\text{STAB}$ (Tesseract 3$^\clubsuit$).}
\end{subfigure}
\begin{subfigure}[!htbp]{\FigWidth}
\includegraphics[width=1.0\columnwidth]{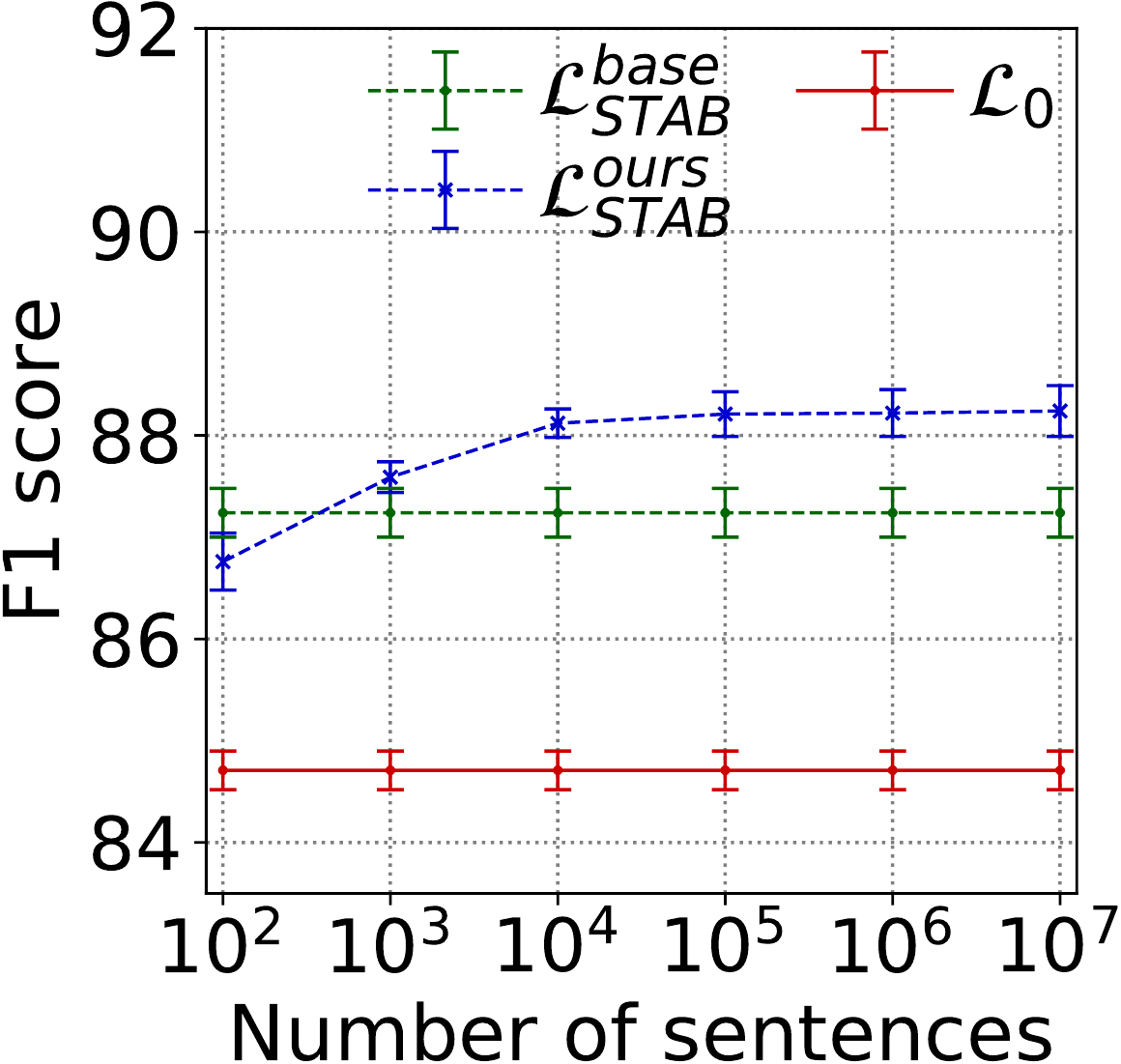}
\caption{$\mathcal{L}_\text{STAB}$ (Tesseract 4$^\diamondsuit$).}
\end{subfigure}
\begin{subfigure}[!htbp]{\FigWidth}
\includegraphics[width=1.0\columnwidth]{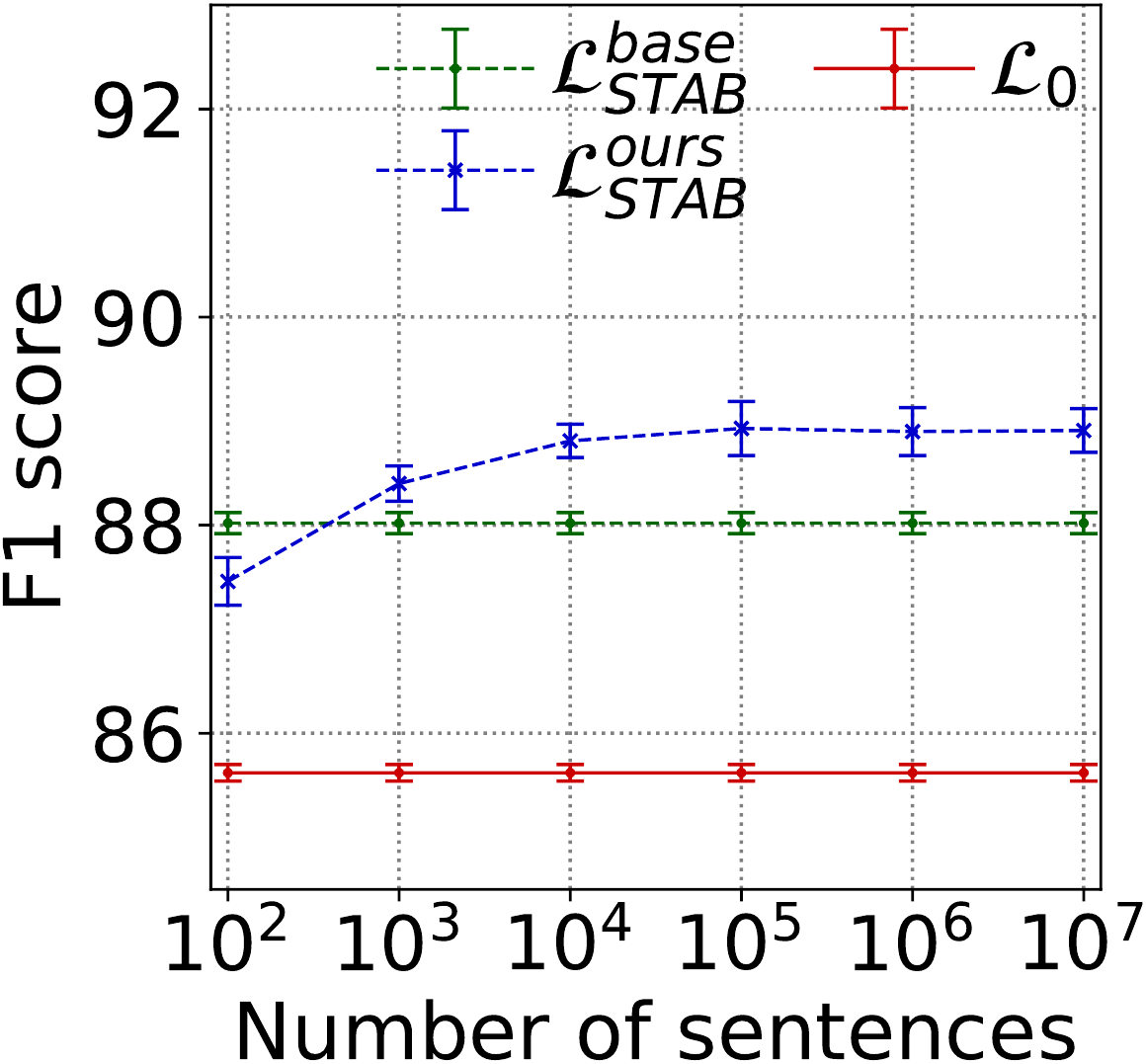}
\caption{$\mathcal{L}_\text{STAB}$ (Tesseract 4$^\heartsuit$).}
\end{subfigure}
\begin{subfigure}[!htbp]{\FigWidth}
\includegraphics[width=1.0\columnwidth]{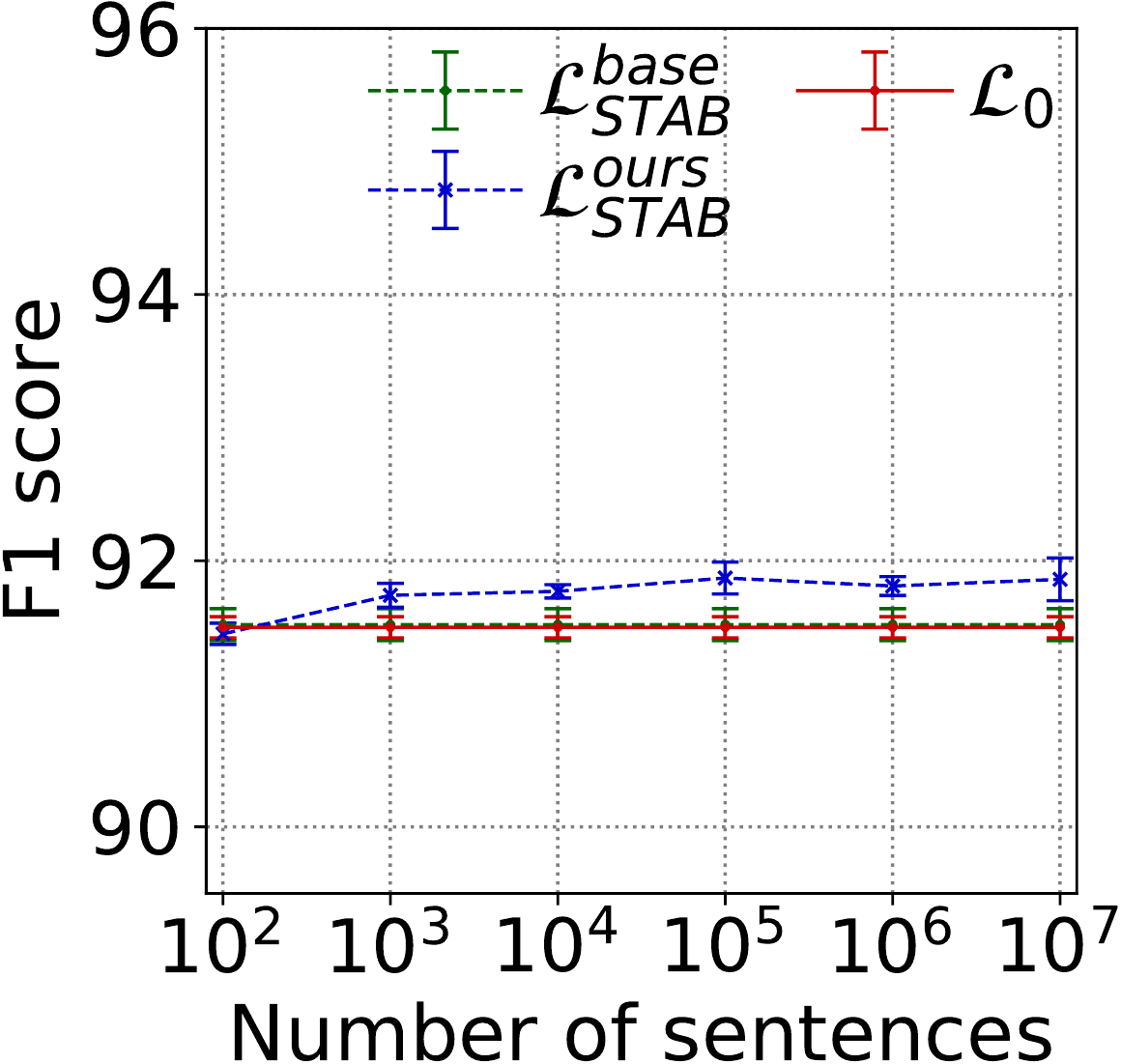}
\caption{$\mathcal{L}_\text{STAB}$ (Tesseract 4$^\spadesuit$).}
\end{subfigure}
\par\smallskip % force a bit of vertical whitespace
\begin{subfigure}[!htbp]{\FigWidth}
\includegraphics[width=1.0\columnwidth]{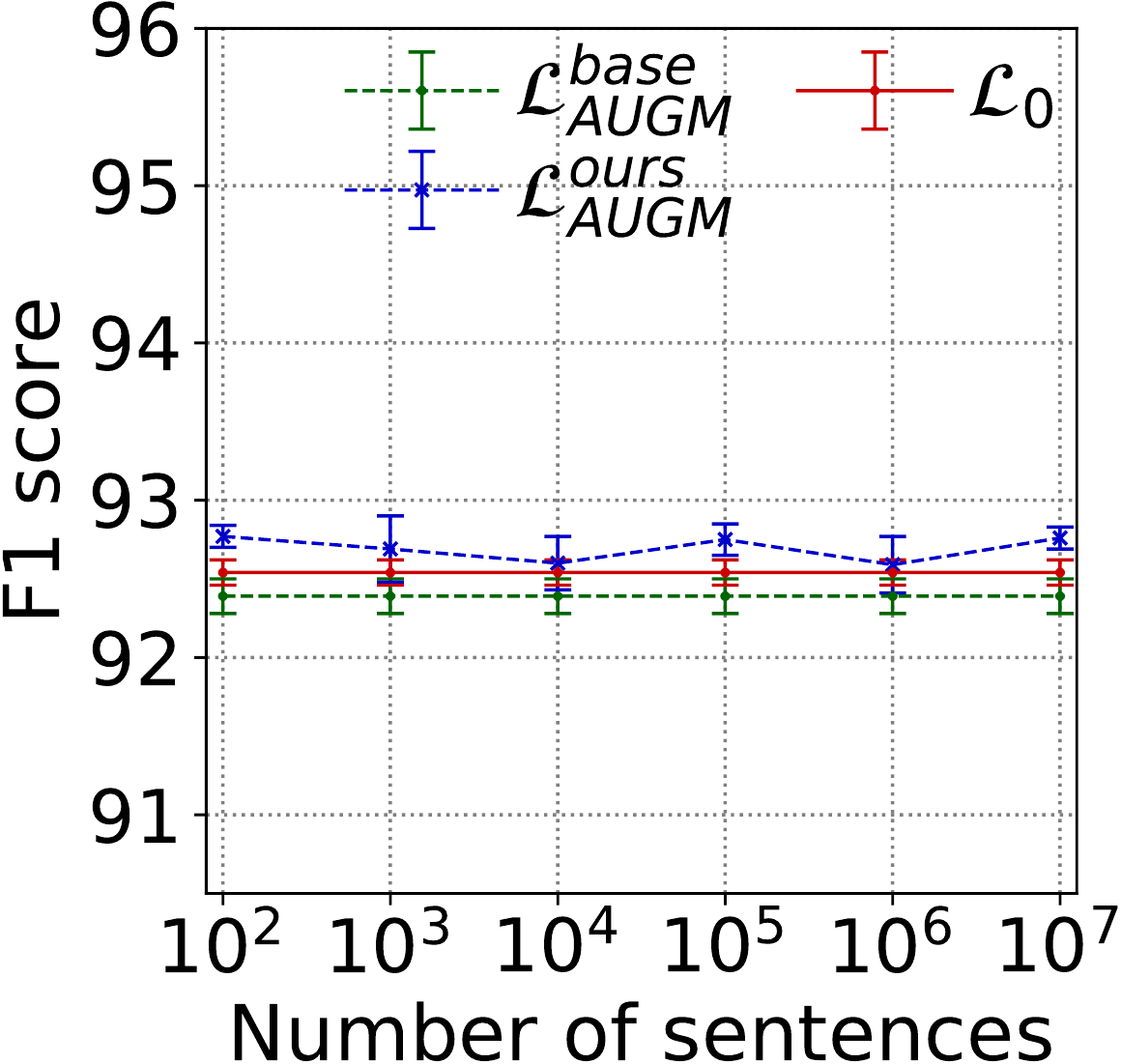}
\caption{$\mathcal{L}_\text{AUGM}$ (original data set).}
\end{subfigure}
\begin{subfigure}[!htbp]{\FigWidth}
\includegraphics[width=1.0\columnwidth]{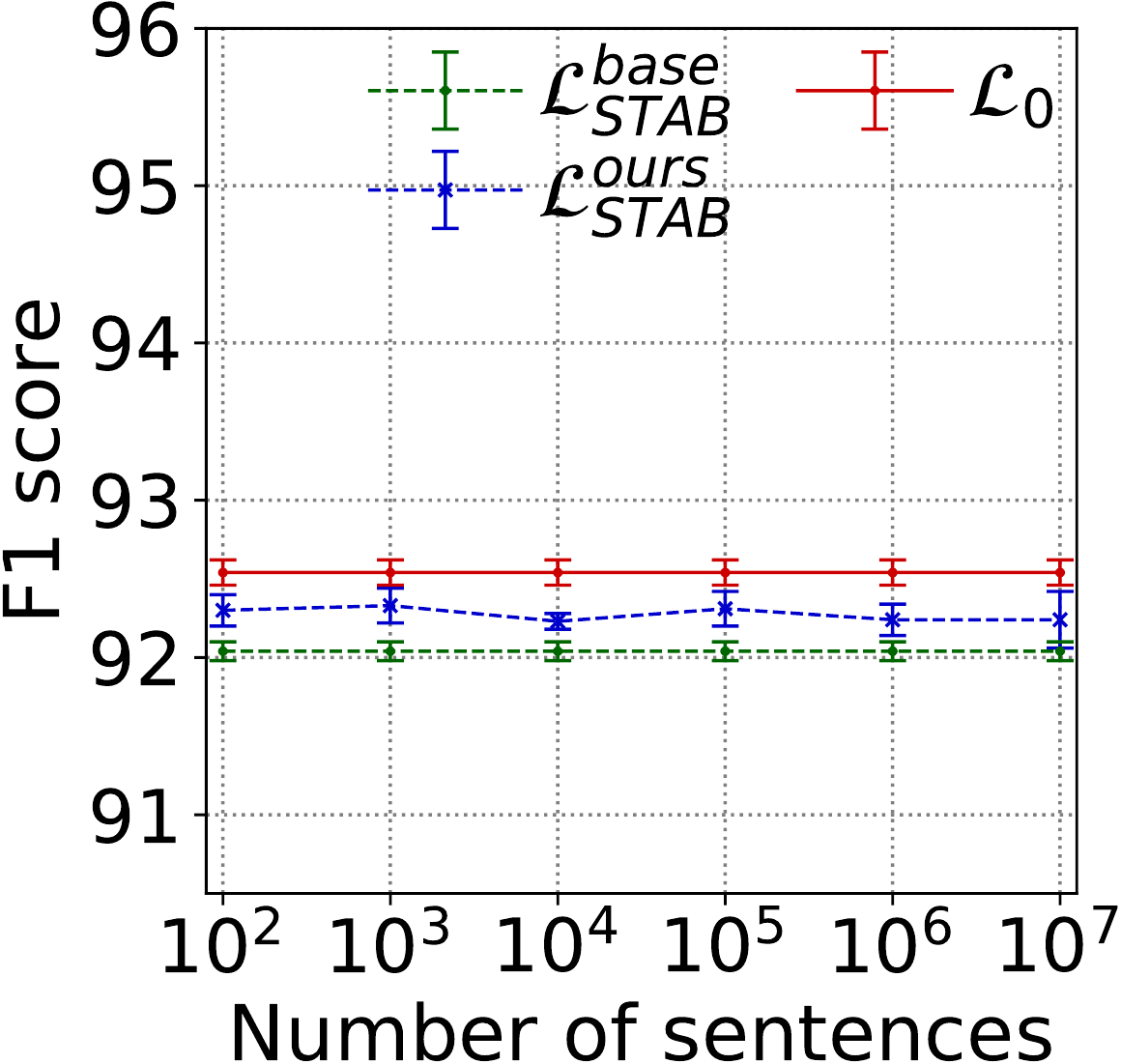}
\caption{$\mathcal{L}_\text{STAB}$ (original data set).}
\end{subfigure}
\begin{subfigure}[!htbp]{\FigWidth}
\includegraphics[width=1.0\columnwidth]{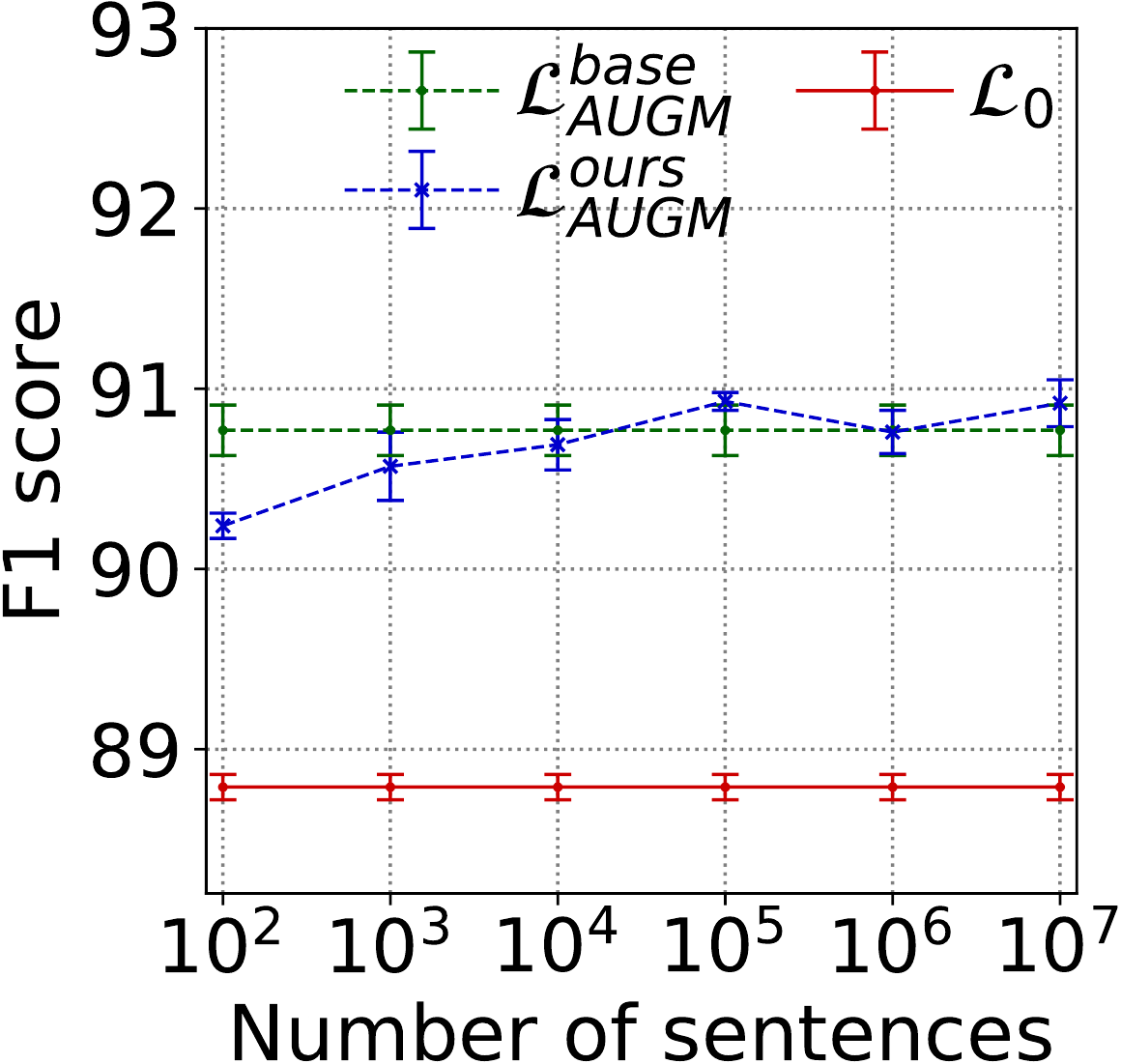}
\caption{$\mathcal{L}_\text{AUGM}$ (Typos).}
\end{subfigure}
\begin{subfigure}[!htbp]{\FigWidth}
\includegraphics[width=1.0\columnwidth]{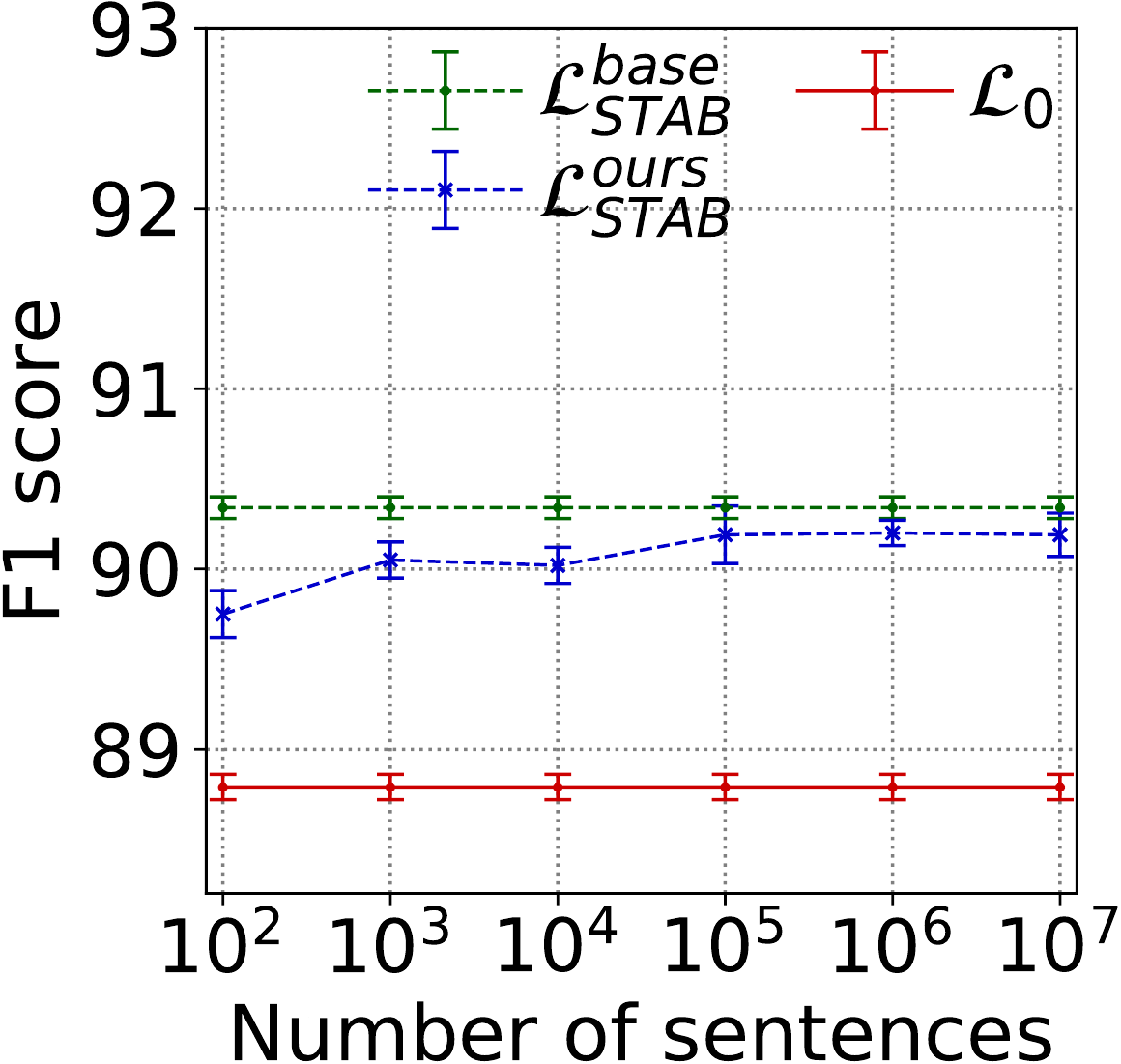}
\caption{$\mathcal{L}_\text{STAB}$ (Typos).}
\end{subfigure}
\caption{F1 score in relation to the number of parallel sentences.
The experiments were conducted on the original English CoNLL 2003 benchmark and its noisy variants: Tesseract 3$^\clubsuit$, Tesseract 4$^\diamondsuit$, Tesseract 4$^\heartsuit$, Tesseract 4$^\spadesuit$, and Typos.
We compare the accuracy of our {\tok} seq2seq approach with the vanilla error model {\mnp}, and the standard objective ($\mathcal{L}_0$).
We present the results for both auxiliary objectives: the data augmentation ($\mathcal{L}_\text{AUGM}^\text{ours}$, $\mathcal{L}_\text{AUGM}^\text{base}$) and the stability training ($\mathcal{L}_\text{STAB}^\text{ours}$, $\mathcal{L}_\text{STAB}^\text{base}$).
}
\label{fig:eval-low-res}
\end{center}
\end{figure*}
}

%==================== Sequence Labeling Data Sets ====================

\section{Sequence Labeling Data Sets}
\label{sec:data_sets}

\paragraph{Original Benchmarks}

\Cref{tab:data_sets} presents the detailed statistics of the original sequence labeling benchmarks used in our experiments.
For NER, we employed CoNLL 2003\footnote{\url{https://www.clips.uantwerpen.be/conll2003/ner}}~\citep{tjong-kim-sang-de-meulder-2003-introduction}.
To evaluate POST, we utilized Universal Dependency Treebank (UD English EWT\footnote{\url{https://universaldependencies.org/treebanks/en_ewt} (version 2.6)};~\citealp{silveira-etal-2014-gold}).

{\setlength{\tabcolsep}{9pt}\renewcommand{\arraystretch}{1.0}
\begin{table}[!htb]
\begin{subtable}{\columnwidth}\centering\small
\begin{tabular}{@{}lrrrr@{}}
\toprule
          & Train & Dev & Test & Total\\
\midrule
Sentences & 14,041  & 3,250  & 3,453  & 20744 \\
Tokens    & 203,621 & 51,362 & 46,435 & 301418 \\
PER       & 6,600   & 1,842  & 1,617  & 10059 \\
LOC       & 7,140   & 1,837  & 1,668  & 10645 \\
ORG       & 6,321   & 1,341  & 1,661  & 9323 \\
MISC      & 3,438   & 922   & 702     & 5062 \\
\bottomrule
\end{tabular}
\caption{English CoNLL 2003.}
\end{subtable}
\par\smallskip % force a bit of vertical whitespace
\begin{subtable}{\columnwidth}\centering\small
\begin{tabular}{@{}lrrrr@{}}
\toprule
          & Train & Dev & Test & Total\\
\midrule
Sentences & 12543  & 2002  & 2077  & 16622  \\
Tokens    & 204585 & 25148 & 25096 & 254829 \\
%\midrule
%%% UD v2.6
ADJ       & 12458  & 1784  & 1689  & 15931 \\
ADP       & 17625  & 2021  & 2020  & 21666 \\
ADV       & 10553  & 1264  & 1226  & 13043 \\
AUX       & 12396  & 1512  & 1504  & 15412 \\ 
CCONJ     & 6703   & 780   & 738   & 8221  \\
DET       & 16284  & 1895  & 1896  & 20075 \\
INTJ      & 688    & 115   & 120   & 923   \\
NOUN      & 34765  & 4196  & 4129  & 43090 \\
NUM       & 3996   & 378   & 536   & 4910  \\
PART      & 5567   & 630   & 630   & 6827  \\
PRON      & 18584  & 2219  & 2158  & 22961 \\
PROPN     & 12945  & 1879  & 2075  & 16899 \\
PUNCT     & 23676  & 3083  & 3106  & 29865 \\
SCONJ     & 3850   & 403   & 386   & 4639  \\
SYM       & 643    & 75    & 100   & 818   \\
VERB      & 23005  & 2759  & 2644  & 28408 \\
X         & 847    & 155   & 139   & 1141  \\
\bottomrule
\end{tabular}
\caption{UD English EWT.}
\end{subtable}
\caption{
Statistics of the English CoNLL 2003 and the UD English EWT data sets. 
We present statistics of the training (Train) development (Dev) and test (Test) sets, including the number of sentences and tokens. 
CoNLL 2003 contains the annotations for the following entity types: person names (PER), locations (LOC), organizations (ORG), and miscellaneous (MISC). 
For UD English EWT, the following universal POS tags were included: ADJ (adjective), ADP (adposition), ADV (adverb), AUX (auxiliary), CCONJ (coordinating conjunction), DET (determiner), INTJ (interjection), NOUN (noun), NUM (numeral), PART (particle), PRON (pronoun), PROPN (proper noun), PUNCT (punctuation), SCONJ (subordinating conjunction), SYM (symbol), VERB (verb), X (other).}
\label{tab:data_sets}
\end{table}}

\paragraph{Noisy Benchmarks}

{\renewcommand{\arraystretch}{0.9}\setlength{\tabcolsep}{3.0pt}
\newcommand{\ColWidthA}{0.14}
\begin{table}[!h]
\begin{subtable}{\columnwidth}\centering\small
\begin{tabular}{@{}X{\ColWidthA}X{0.14}*{5}{Z{0.11}}@{}}
\toprule
Measure & Method & Tesser-act~3$^\clubsuit$ & Tesser-act~4$^\diamondsuit$ & Tesser-act~4$^\heartsuit$ & Tesser-act~4$^\spadesuit$ & Typos\\
\midrule
\multirow{3}{\ColWidthA\columnwidth}{\centering TER}
& Original & 22.72 & 16.35 & 14.89 & 3.53 & 15.53\\
& Natas    & {\bf17.24} & {\bf12.20} & {\bf11.13} & {\bf2.34} & 11.53 \\
& Hunspell & 17.44 & 13.54 & 12.24 & 2.43 & {\bf10.69} \\
\midrule
\multirow{3}{\ColWidthA\columnwidth}{\centering TER (entities)}
& Original      & 29.66 & 16.70 & 15.00 & 3.61 & 8.20 \\
& Natas    & 27.81 & 14.97 & 13.36 & 2.93 & 7.62 \\
& Hunspell & {\bf16.63} & {\bf9.95}  & {\bf8.76} & {\bf1.89} & {\bf4.07} \\
\midrule
\multirow{2}{\ColWidthA\columnwidth}{\centering ACC}
& Natas    & {\bf24.13} & {\bf25.40} & {\bf25.24} & {\bf33.70} & 25.75\\
& Hunspell & 23.26 & 17.19 & 17.76 & 31.20 & {\bf31.17} \\
\midrule
\multirow{2}{\ColWidthA\columnwidth}{\centering ACC (entities)}
& Natas    & 6.23  & 10.41 & 10.93 & 18.77 & 7.07\\
& Hunspell & {\bf43.93} & {\bf40.44} & {\bf41.58} & {\bf47.78} & {\bf50.38} \\
\bottomrule
\end{tabular}
\caption{English CoNLL 2003.}
\end{subtable}
\par\smallskip % force a bit of vertical whitespace
\begin{subtable}{\columnwidth}\centering\small
\begin{tabular}{@{}X{\ColWidthA}X{0.14}*{5}{Z{0.11}}@{}}
\toprule
Measure & Method & Tesser-act~3$^\clubsuit$ & Tesser-act~4$^\diamondsuit$ & Tesser-act~4$^\heartsuit$ & Tesser-act~4$^\spadesuit$ & Typos\\
\midrule
\multirow{3}{\ColWidthA\columnwidth}{\centering TER}
& Original & 23.31 & 22.12 & 20.38 & 5.83 & 15.22 \\
& Natas    & {\bf17.76} & {\bf17.46} & {\bf16.23} & {\bf4.21} & 11.68 \\
& Hunspell & 19.14 & 19.74 & 18.09 & 4.75 & {\bf11.22} \\
\midrule
\multirow{2}{\ColWidthA\columnwidth}{\centering ACC}
& Natas    & {\bf23.82} & {\bf21.05} & {\bf20.36} & {\bf27.75} & 23.27\\
& Hunspell & {17.90} & {10.74} & {11.20} & {18.59} & {\bf26.49} \\
\bottomrule
\end{tabular}
\caption{UD English EWT.}
\end{subtable}
\caption{Token Error Rates (TER) and the correction accuracies (ACC) of Natas and Hunspell on the test sets of our noisy sequence labeling data sets.
All values are percentages.
{\bf Bold} values represent the lowest TER and the highest ACC.
}
\label{tab:error_rates}
\end{table}}

\Cref{tab:error_rates} presents the error rates and the correction accuracies of the Natas and Hunspell methods calculated on the test sets of the noisy sequence labeling benchmarks.
Moreover, \Cref{tab:conll-format} shows an excerpt from a noisy sequence labeling data set generated for evaluation.
{\renewcommand{\arraystretch}{1.0}
\begin{table}[htbp]
\centering\small
\begin{tabular}{lll}
\toprule
Noisy Token & Error-Free Token & Class Label\\
\midrule
\verb|No| & \verb|No| & \verb|O|\\
\verb|nzw| & \verb|new| & \verb|O|\\
\verb|fixtuvzs| & \verb|fixtures| & \verb|O|\\
\verb|reported| & \verb|reported| & \verb|O|\\
\verb|from| & \verb|from| & \verb|O|\\
\verb|New| & \verb|New| & \verb|B-LOC|\\
\verb|Vork| & \verb|York| & \verb|I-LOC|\\
\verb|.| & \verb|.| & \verb|O|\\
\bottomrule
\end{tabular}
\caption{Example of a sentence from the noisy CoNLL 2003 data set.
The first and the second column contains the noisy and the error-free tokens, respectively.
The third column denotes the class label in BIO format. 
}
\label{tab:conll-format}
\end{table}}

\begin{figure*}[!htb]
\begin{center}
\begin{subfigure}[!htbp]{0.48\textwidth}
\includegraphics[width=1.0\columnwidth]{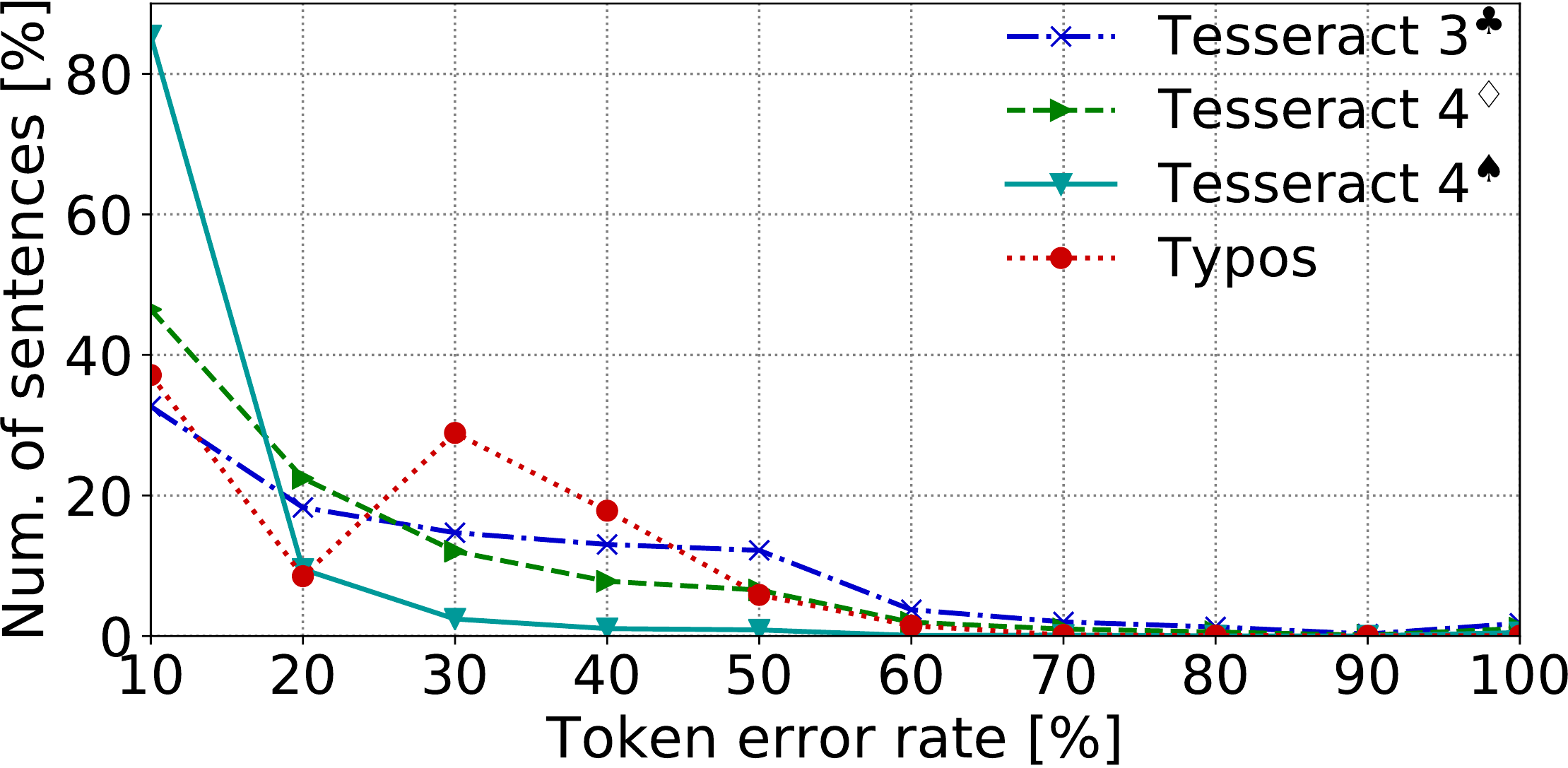}
\caption{Noisy sentence labeling data sets (English CoNLL 2003).}
\end{subfigure}
\quad
\begin{subfigure}[!htbp]{0.48\textwidth}
\includegraphics[width=1.0\columnwidth]{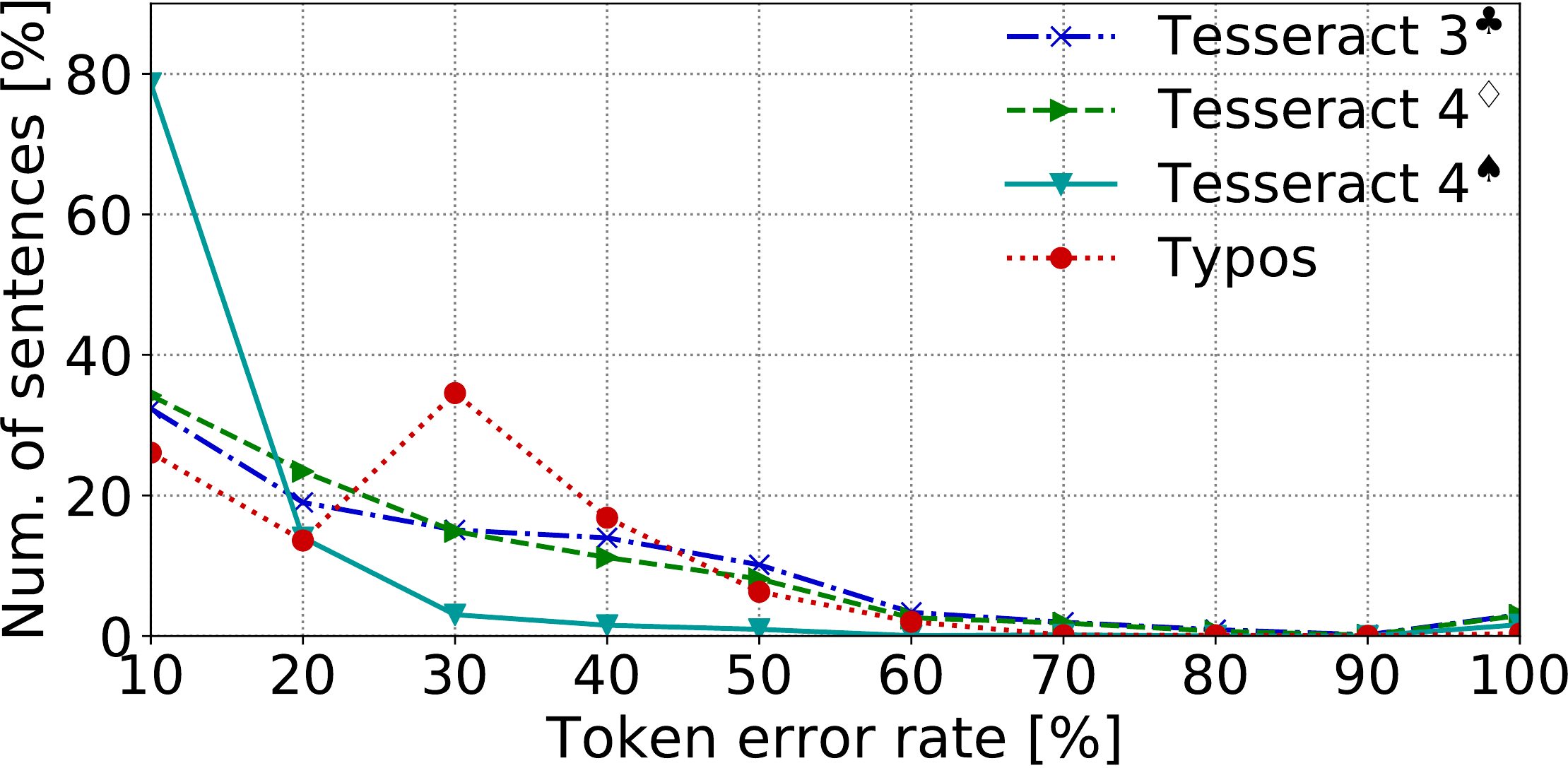}
\caption{Noisy sentence labeling data sets (UD English EWT).}
\end{subfigure}
\par\smallskip % force a bit of vertical whitespace
\par\smallskip % force a bit of vertical whitespace
\begin{subfigure}[!htb]{0.48\textwidth}
\includegraphics[width=1.0\columnwidth]{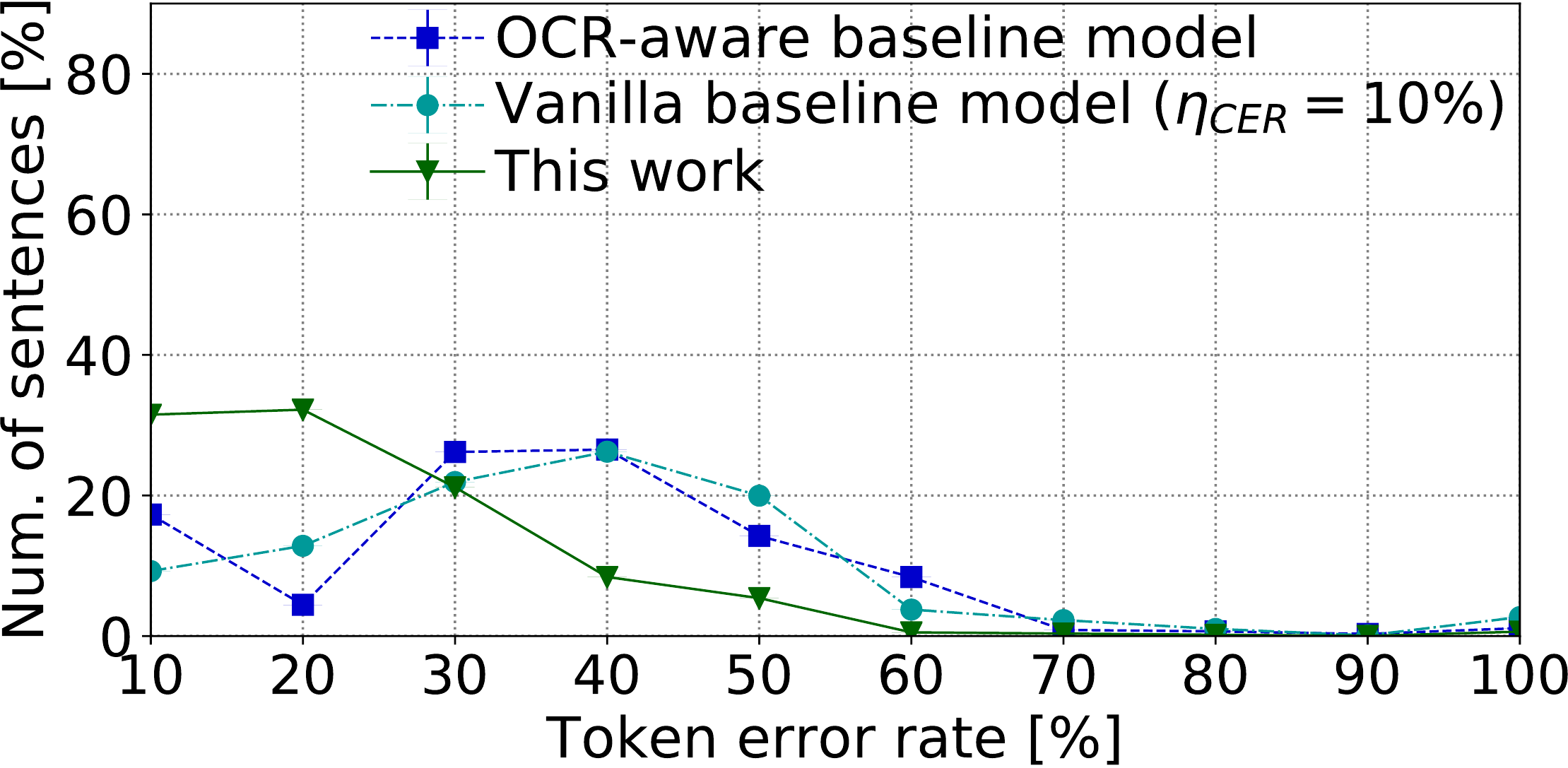}
\caption{Noise induction methods (English CoNLL 2003).}
\end{subfigure}
\quad
\begin{subfigure}[!htbp]{0.48\textwidth}
\includegraphics[width=1.0\columnwidth]{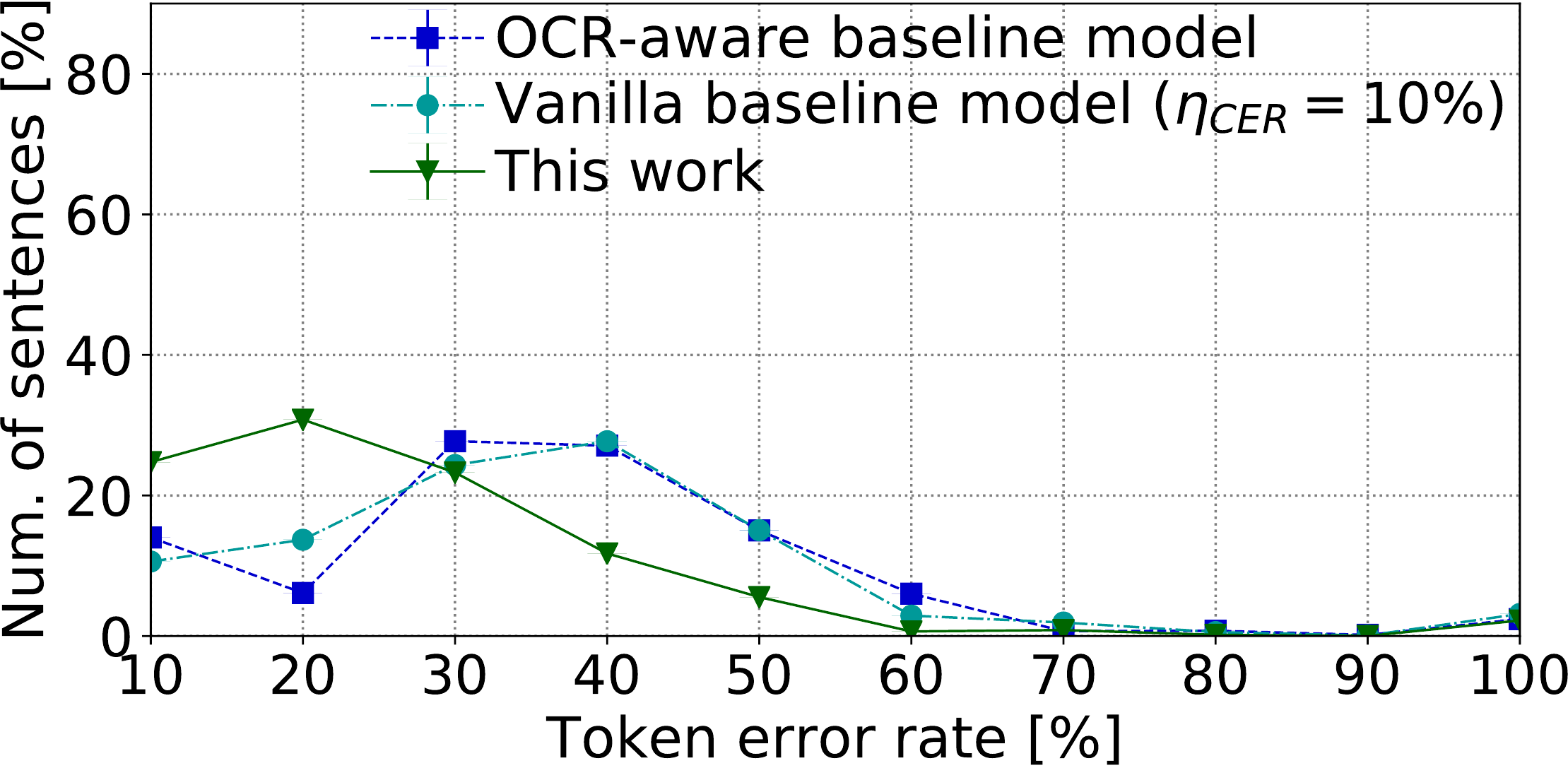}
\caption{Noise induction methods (UD English EWT).}
\end{subfigure}
\caption{
\narrowstyleC
{
Distributions of the token error rates of sentences in our noisy sequence labeling data sets (Tesseract 3$^\clubsuit$, Tesseract 4$^\diamondsuit$, Tesseract 4$^\spadesuit$, and Typos).
For comparison, we include error distributions obtained by applying our seq2seq {\tok} error generator and the baseline confusion matrix-based error models {\mnp} to the sentences extracted from the original benchmark.
$\eta_\text{CER}$ is the character-level noising factor used by the vanilla error model.
Each point is the percentage of sentences with a token error rate that falls into a specific token error range, i.e., the value of 50 corresponds to the sentences with a token error rate greater than 40 and lower than or equal to 50.
}
}
\label{fig:error-distrib2}
\end{center}
\end{figure*}

Furthermore, \Cref{fig:error-distrib2} presents the distribution of token error rates in relation to the percentage number of sentences in our noisy data sets.
For comparison, we also included the distributions obtained by applying different noise generation methods - the vanilla- and the OCR-aware confusion matrix-based models by {\mnt}, and our {\tok} seq2seq error generator.

We note that the error distribution of our noisy data sets is closer to the Zipf distribution in contrast to the results of prior methods that exhibit a Bell-Curve pattern.
Note that the \emph{Typos} data set was generated by randomly sampling possible lexical replacement candidates from the lookup tables, hence its distribution exhibits slightly different characteristics than the noisy data sets generated by directly applying the OCR engine to the rendered text images.
Based on the above results, we believe that our noisy data sets are better suited for the evaluation of the robustness of sequence labeling models than the data generated by the prior approaches. 

\paragraph{Data Conversion Scripts}
Because of licensing and copyright reasons, we did not submit the noisy data sets directly. 
Our code includes the scripts for the conversion of the original benchmarks into their noisy variants.
For reference, we added excerpts of the noisy UD English EWT data set in the supplementary materials.

%

%==================== Reproducibility ====================

\section{Reproducibility}
\label{sec:repro}

In this section, we present additional information that could facilitate reproducibility.

\paragraph{Hyper-parameters}

To train our seq2seq translation models, we generally used the default hyper-parameters of the OpenNMT toolkit.
We list all non-default values in \Cref{tab:onmt-hyperparams}.
Moreover, we decayed the learning rate eight times during the training for all models.
Furthermore, we utilized \textit{copy attention}~\citep{see-etal-2017-get} for our error generation models and \textit{global attention}~\citep{luong-etal-2015-effective} for the error correction model.

{\setlength{\tabcolsep}{3pt}\renewcommand{\arraystretch}{1.0}
\newcommand{\ColWidthA}{0.42}
\newcommand{\ColWidthB}{0.42}
\newcommand{\ColWidthC}{0.1}
\begin{table}[!h]\centering\small
\begin{tabular}{@{}X{\ColWidthA}X{\ColWidthB}Z{\ColWidthC}@{}}
\toprule
Parameter & Description & Value \\
\midrule
\verb|-share_vocab| & Share source and target vocabulary &  True \\
\addlinespace[0.15cm]
\verb|-share_embeddings| & Share the embeddings between encoder and decoder & True \\
\addlinespace[0.15cm]
\verb|-word_vec_size| & Word embedding size & 25 \\
\addlinespace[0.15cm]
\verb|-src_seq_length| & Max. source sequence length & 1000 \\
\addlinespace[0.15cm]
\verb|-tgt_seq_length| & Max. target sequence length & 1000 \\
\addlinespace[0.15cm]
\verb|-encoder_type| & Type of encoder layer & brnn \\
\addlinespace[0.15cm]
\verb|-learning_rate| & Starting learning rate & 1.0 \\
\bottomrule
\end{tabular}
\caption{
\narrowstyleB
{
The hyper-parameters of the OpenNMT toolkit used to train our seq2seq error generation models.
}
}
\label{tab:onmt-hyperparams}
\end{table}}

\paragraph{Validation Accuracy}
\Cref{tab:valid-acc} summarizes the validation accuracy of our seq2seq models for error generation.
We trained the {\ch} models for $1.6\times10^4$ and the {\tok} models for $4\times10^5$ iterations or at least one epoch of training.
Moreover, the {\tok} error correction model employed by Natas was trained for one epoch (about $4\times10^5$ iterations) on one million parallel sentences and achieved $96.9\%$ accuracy on the validation set of $5000$ sentences.

{\setlength{\tabcolsep}{5.5pt}\renewcommand{\arraystretch}{1.0}
\newcommand{\ColWidth}{0.18}
\begin{table}[!h]\centering\small
\begin{tabular}{@{}Z{\ColWidth}Z{\ColWidth}Z{0.24}Z{0.24}@{}}
\toprule
\multirow{2}{\ColWidth\columnwidth}[-4pt]{\centering Training set size} & \multirow{2}{\ColWidth\columnwidth}[-4pt]{\centering Validation set size} & \multicolumn{2}{c}{Validation accuracy} \\
\cmidrule{3-4}
& & {\tok} & {\ch} \\
\midrule
$10^7$ & 5000 & 98.3\% & 95.7\% \\
$10^6$ & 5000 & 95.4\% & 94.9\% \\
$10^5$ & 5000 & 95.1\% & 95.3\% \\
$10^4$ & 1000 & 94.6\% & 90.1\% \\
$10^3$ & 100  & 93.3\% & 91.6\% \\
\bottomrule
\end{tabular}
\caption{
\narrowstyleC
{
Validation accuracy of the seq2seq models for error generation.
We trained both the {\tok} and the {\ch} variants.
The first and the second column show the number of parallel sentences used for training and validation, respectively.
}
}
\label{tab:valid-acc}
\end{table}}

\paragraph{Learnable Parameters}
The number of parameters in our sequence labeling models was constant among different models, as we used the same architecture in all experiments.
The number of all model parameters was 60.3 million (including embeddings that were fixed during the training), and the number of all trainable parameters was 25.5 million. 
Moreover, all our seq2seq error generation and correction models had about 7.7 million parameters.

\paragraph{Average Runtime}
The evaluation of the complete test set took 7 and 10 seconds on average in the case of UD English EWT and English CoNLL 2003, respectively. 
The runtime did not depend on the training method that was used.
Nevertheless, when we employed the correction method, the runtime was significantly lengthened, e.g., it took almost 3 minutes to evaluate a model that employed the Natas correction method on English CoNLL 2003.

\paragraph{Computing Architecture}
The evaluation was performed on a workstation equipped with an Intel Xeon CPU with 10 cores and an Nvidia Quadro RTX 6000 graphics card with 24GB of memory.

\onlyinsubfile{
\bibliographystyle{../acl_natbib}
\bibliography{../anthology,../acl2021}
}